\documentclass[journal]{IEEEtran}

\usepackage{times}
\usepackage{epsfig}

\usepackage{graphics} %
\usepackage{graphicx}
\usepackage{grffile}
\usepackage{algpseudocode,algorithm}

\usepackage{amsmath,amsfonts,amssymb}
\usepackage{color}
\usepackage{wrapfig}
\usepackage{subfigure}
\usepackage{multirow}
\usepackage{array}
\usepackage[table]{xcolor}
\usepackage{soul}
\usepackage{mathtools}
\usepackage{fixmath}
\usepackage{nccmath}
\usepackage{cuted}
\usepackage{float}
\usepackage{caption}
\usepackage{siunitx}
\usepackage{url}
\usepackage{graphicx}
\usepackage{booktabs}
\usepackage{makecell}
\usepackage{enumitem} %
\usepackage{adjustbox}
\newcolumntype{P}[1]{>{\centering\arraybackslash}p{#1}}
\newcolumntype{M}[1]{>{\centering\arraybackslash}m{#1}}
\usepackage{cite} %

\newcommand{\ie}{\textit{i}.\textit{e}.}
\newcommand{\eg}{\textit{e}.\textit{g}.}

\newcolumntype{L}[1]{>{\raggedright\let\newline\\\arraybackslash\hspace{0pt}}m{#1}}
\newcolumntype{C}[1]{>{\centering\let\newline\\\arraybackslash\hspace{0pt}}m{#1}}
\newcolumntype{R}[1]{>{\raggedleft\let\newline\\\arraybackslash\hspace{0pt}}m{#1}}

\usepackage{threeparttable}

\global\long\def\bx{\mathbf{x}}

\global\long\def\cS{\mathcal{S}} %

\global\long\def\bV{\mathbf{V}}

\global\long\def\siOmega{\SI{}{\Omega}}
\global\long\def\cM{\mathcal{M}}
\global\long\def\cL{\mathcal{L}}
\global\long\def\Lum{L}
\global\long\def\pol{p}

\global\long\def\Warp{\mathbf{W}} %
\global\long\def\bm{\mathbf{m}} %
\global\long\def\cE{\mathcal{E}} %
\global\long\def\cS{\mathcal{S}} %

\def\MYTITLE{Event-based Motion Segmentation with Spatio-Temporal Graph Cuts}

\newcommand\MYhyperrefoptions{bookmarks=true,bookmarksnumbered=true,
pdfpagemode={UseOutlines},plainpages=false,pdfpagelabels=true,
colorlinks=true,breaklinks=true,
pdftitle={\MYTITLE},%
pdfsubject={Robotics, Computer Vision},%
pdfauthor={Yi Zhou, Guillermo Gallego, Xiuyuan Lu, Siqi Liu, Shaojie Shen},%
pdfkeywords={Event camera, Segmentation, Motion compensation, Graph cut, Asynchronous sensor}}%
\usepackage[\MYhyperrefoptions,pdftex]{hyperref}

\ifCLASSINFOpdf
\else
\fi

\title{\MYTITLE}

\begin{document}

\newif\iflongversion

\author{Yi Zhou, Guillermo Gallego, Xiuyuan Lu, Siqi Liu, Shaojie Shen
\thanks{Y. Zhou is with School of Robotics at Hunan University, Changsha, China.
G. Gallego is with the Technische Universit\"at Berlin and the Einstein Center Digital Future, Berlin, Germany.
X. Lu, S. Liu and S. Shen are with Robotic Institute, 
the Department of Electronic and Computer Engineering at the Hong Kong University of Science and Technology, Hong Kong, China. 
E-mail: eeyzhou@hnu.edu.cn, eeshaojie@ust.hk. 
}
\thanks{This work was supported by the HKUST Institutional Fund and the HKUST Postdoc Fellowship Matching Fund 2020. (Corresponding author: Yi Zhou.)}
}

\markboth{IEEE Transactions on Neural Networks and Learning Systems. PREPRINT VERSION. ACCEPTED Oct.~2021.}%
{}

\maketitle

\begin{abstract}
Identifying independently moving objects is an essential task for dynamic scene understanding.
However, traditional cameras used in dynamic scenes may suffer from motion blur or exposure artifacts due to their sampling principle. 
By contrast, event-based cameras are novel bio-inspired sensors that offer advantages to overcome such limitations. 
They report pixel-wise intensity changes asynchronously, which enables them to acquire visual information at exactly the same rate as the scene dynamics.
We develop a method to identify independently moving objects acquired with an event-based camera, i.e., to solve the event-based motion segmentation problem.
We cast the problem as an energy minimization one involving the fitting of multiple motion models.
We jointly solve two subproblems, namely event-cluster assignment (labeling) and motion model fitting, in an iterative manner by exploiting the structure of the input event data in the form of a spatio-temporal graph.
Experiments on available datasets demonstrate the versatility of the method in scenes with different motion patterns and number of moving objects.
The evaluation shows state-of-the-art results without having to predetermine the number of expected moving objects.
We release the software and dataset under an open source licence to foster research in the emerging topic of event-based motion segmentation.
\end{abstract}

\begin{IEEEkeywords}
Event-based Vision, Motion Segmentation, Motion Compensation, Graph Cut.
\end{IEEEkeywords}

\section*{Multimedia Material}
\noindent Project page: {\small \url{https://sites.google.com/view/emsgc}}\\
Code: {\small \url{https://github.com/HKUST-Aerial-Robotics/EMSGC.git}}\\

\section{Introduction}
\label{sec: introduction}
\begin{figure*}
  \centering
  \includegraphics[trim={.0cm 0.04cm 0.cm 0.cm},clip,width=\linewidth]{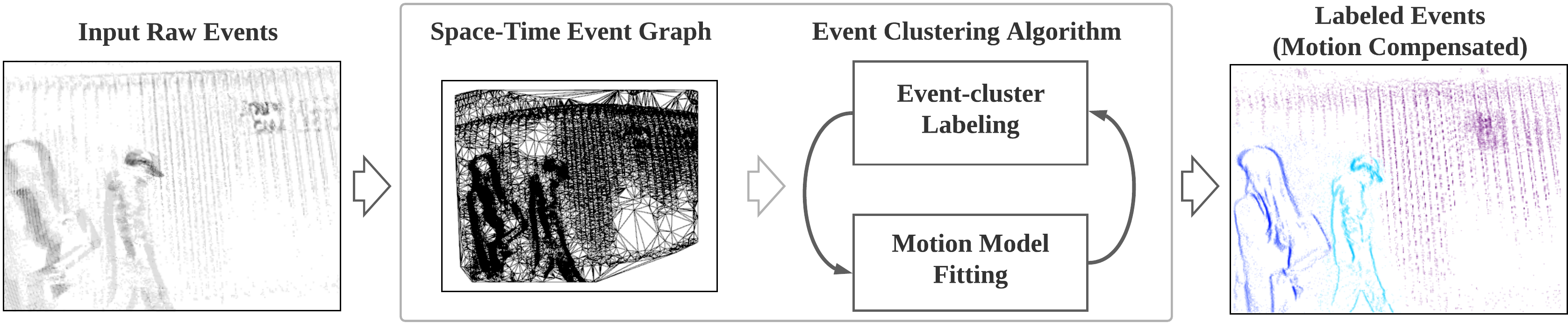}
  \captionof{figure}{\emph{Input / Output}: The proposed method classifies the events acquired by an event-based camera (Left: events collapsed along time) into different groups that undergo independent motions (Right). %
  \emph{Approach} (Middle block):
  We create a space-time event graph %
  and pass it to an iterative clustering algorithm that jointly classifies events into objects and fits motion models to them.
  The method fits the event data globally while encouraging spatial coherence and fewest number of clusters.
  }
  \label{fig:eyecatcher}
\end{figure*}

Event-based cameras, such as the Dynamic Vision Sensor (DVS) \cite{Lichtsteiner08ssc,Suh20iscas,Finateu20isscc}, are novel bio-inspired visual sensors.
Unlike standard cameras that acquire data at a fixed rate, event-based cameras report per-pixel intensity changes asynchronously at the time they occur, with microsecond resolution, called ``events''.
This working principle offers potential advantages (low latency, high temporal resolution, high dynamic range and low power consumption) to tackle challenging scenarios in computer vision such as high-speed and/or high dynamic range (HDR) optical flow estimation~\cite{Benosman14tnnls,Zhu18rss}, feature tracking~\cite{Lagorce15tnnls,Valeiras15tnnls,Zhu17icra,Gehrig19ijcv},
stereo depth estimation~\cite{Rogister12tnnls,Lee14tnnls,Osswald17srep,Zhou18eccv},
camera tracking~\cite{Mueggler15rss,Gallego17pami,Chamorro20bmvc},
control~\cite{Conradt09iscas,Delbruck13fns} and Simultaneous Localization and Mapping (SLAM)~\cite{Kim16eccv,Rebecq17ral,Rosinol18ral,Mueggler18tro,zhou2020event}.
However, due to the above principle of operation and unconventional output, algorithms designed for standard cameras cannot be directly applied.
Novel algorithms are needed to unlock event cameras' potential.
The recent survey \cite{Gallego20pami} provides a comprehensive review on event-based cameras, algorithms and applications.

In this paper, we consider the problem of event-based motion segmentation, which aims at classifying events occurred during a time interval into several groups that represent coherent moving objects.
We tackle the most general case: a possibly moving event camera observing a dynamic scene.
It is more challenging than the static camera case because events are no longer solely due to IMOs; they are also induced by the background.
We develop a method to jointly classify events and estimate their coherent motion, in an iterative and alternating manner.

\textbf{Contributions}. Our contributions can be summarized as:
\begin{enumerate}
    \item A novel event-based motion segmentation method designed in the spirit of motion compensation~\cite{Gallego18cvpr} and built on top of classical multi-model fitting schemes.
    We propose a space-time event graph representation to exploit the spatio-temporal nature of events, leading to globally consistent and locally coherent labeling results.
    
    \item A simple and effective initialization method using the original motion compensation scheme on raw events.
    
    \item A formulation that does not require prior knowledge in the form of scene geometry, motion patterns or number of IMOs.%
    
    \item An extensive evaluation, qualitative and quantitative, on %
    available datasets, showing better performance than directly competing baseline methods.
\end{enumerate}
\vspace{1ex}
The rest of the paper is organized as follows:
Section~\ref{sec:DVS_operation_application} briefly describes the working principle of event-based cameras.
Section~\ref{sec:problem:intro} discusses the nature of the event-based motion segmentation problem and the related work.
Section~\ref{sec:problem-statement} more formally states the problem and necessary preliminaries.
Sections~\ref{sec:energy-formualtion} and~\ref{sec:optimization} disclose our energy formulation and optimization procedure, respectively.
The method is evaluated in Section~\ref{sec:evaluation} and conclusions are drawn in Section~\ref{sec:conclusion}.

\section{Event-based Motion Segmentation}

\subsection{Event-based Camera Working Principle}
\label{sec:DVS_operation_application}

Event-based cameras~\cite{Lichtsteiner08ssc} have independent pixels that continuously monitor their incident light and respond to changes of predefined size $C$, which are called ``events''. 
Specifically, if $\Lum(\bx,t)\doteq \log I(\bx,t)$ is the logarithmic brightness at pixel $\bx \doteq (x,y)^\top$ on the image plane, an event $e_k \doteq (\bx_k,t_k,\pol_k)$ is generated at pixel $\bx_k$ and time $t_k$ (with microsecond resolution) if the change in log-brightness reaches $C$ (typically 10-15\% relative change):
\begin{equation}
\label{eq:generativeEventCondition}
\Delta \Lum \doteq \Lum(\bx_k,t_k) - \Lum(\bx_k,t_k-\Delta t_k) = \pol_{k}\, C,
\end{equation}
where $\Delta t_k$ is the time since the last event at the same pixel $\bx_k$ 
and $\pol_k \in \{+1,-1\}$ is the event polarity (i.e., sign of $\Delta L$).

Hence, each pixel has its own sampling rate (which depends on the visual input) 
and outputs data proportionally to the amount of motion or illumination variations in the scene. 
An event-based camera does not produce images at a constant rate, but rather a stream of \emph{asynchronous}, sparse events in space-time domain.

\subsection{The Clustering Nature of the Problem}
\label{sec:problem:intro}
Assuming constant illumination, events are due to the relative motion between the camera and the scene,
and since events represent brightness changes~\eqref{eq:generativeEventCondition}, they are mostly produced by scene edges (contours, texture, etc.).
If the camera is stationary, events are solely due to moving objects, whereas if the camera moves, events are due to both: 
edges of moving objects and moving edges induced by the camera's ego-motion.

The problem of \emph{event-based motion segmentation} consists of identifying which events in a given event stream are triggered by the same scene object,
and because events are caused by motion, the identification criterion primarily refers to grouping (i.e., \emph{classifying}) events by the type of motion.
Hence it is key to realize that event-based motion segmentation is a \emph{clustering problem by nature}:
splitting the event stream into different groups of events, with each one representing a coherent motion (classification criterion), 
as shown in the input-output of Fig.~\ref{fig:eyecatcher}.
The problem is most challenging in the moving-camera scenario because events are triggered everywhere on the image plane, not just around the moving objects.
As we review next (Section~\ref{sec: related work}), several solution methods have been proposed, and due to the above nature of the problem, they all inherently perform some sort of clustering. 
However, they differ in the clustering technique developed.

\subsection{Related Work}
\label{sec: related work}

Early works on event-based motion segmentation required prior knowledge on either the shape of IMOs (\eg, a circle \cite{Glover16iros}) or the correlation between the tracked geometric primitives and the motion of the event camera~\cite{Vasco17icar}.
Such prior knowledge is no longer required in recent works \cite{Mitrokhin18iros,Stoffregen17acra,Stoffregen19iccv,parameshwara2020moms,Mitrokhin19iros}.
Several of these pipelines follow a sequential strategy, which consists of analyzing the dominant events (\eg, background), then removing these (by empirical thresholding \cite{Mitrokhin18iros,Stoffregen17acra}) and analyzing the remaining events (\ie, the IMOs), \emph{greedily}.

Instead, using \cite{Stoffregen17acra} as initialization, \cite{Stoffregen19iccv} was the first to tackle the problem \emph{jointly}, analyzing all events while solving the two sub-problems of clustering: event-object association (classification) and object / cluster refinement (model-fitting).
By leveraging the idea of motion compensation~\cite{Gallego18cvpr}, it formulated the segmentation problem using an expectation-maximization (EM) approach, which iteratively updated the soft event-cluster associations and the motion model parameters.
It provided per-event segmentation rather than classical bounding-box results.
Recently, \cite{parameshwara2020moms} proposed a similar joint optimization method, but with differences: 
($i$) initialization of IMO models was based on $K$-means clustering of event-based feature tracks, 
and ($ii$) event-cluster assignments were based on morphological operations via empirical thresholding.

In addition to the above methods, \cite{Mitrokhin19iros} proposed a supervised end-to-end learning-based pipeline that simultaneously solved for optical flow, 3D motion and object segmentation.
Although \cite{Mitrokhin19iros} is not closely related to the above approaches (neither is to ours), it is mentioned because it provides a state-of-the-art dataset for segmentation evaluation (Section~\ref{sec:evaluation}).

\paragraph*{Similarities and Differences with Prior Work}
Like previous methods (\eg, \cite{Stoffregen19iccv}), our method also allows general parametric warps (motion models) and performs per-event segmentation.
Besides, we are also able to produce sharp, motion-compensated images as a by-product (Fig.~\ref{fig:eyecatcher}, Right).
However, we claim the following fundamental \emph{differences} compared to previous approaches.
First, we formulate the problem using Markov Random Fields (MRF) and defining a spatio-temporal graph through the events. 
This leads us to efficiently solve the problem using \emph{graph cuts}, which is the first time that they are adapted to work on event data.
Second, we pose the problem as a joint optimization over the motion parameters and event associations, but in contrast to \cite{Stoffregen19iccv} we introduce two spatial regularizers.
In particular, we explicitly minimize the number of clusters and smooth their shape, which is pursued naturally via an energy formulation.
This allows us to solve the issue of not knowing the number of moving objects in the scene~\cite{Stoffregen19iccv}.
Third, digging into details of the event alignment data-fidelity terms, \cite{Stoffregen19iccv} is based on variance maximization, whereas graph cuts require a minimization formulation with non-negative terms. 
For this non-trivial adaptation we build on our work~\cite{zhou2020event} and propose negative Images of Warped Events (IWEs), which have not been investigated for clustering.
Finally, we do not follow the greedy strategy nor do we need additional methods for initialization (\eg, feature tracking~\cite{parameshwara2020moms}).
We provide a new initialization method, based on a hierarchical subdivision of the volume of events to provide a pool of motion instances.

\section{Problem Statement Preliminaries}
\label{sec:problem-statement}

The goal of this work is to cluster the events produced by an event-based camera into groups that undergo coherent motions.
Such motions aim to represent the unknown number of IMOs in the scene.
Once clustered, events are warped according to the estimated motions and produce an overall %
IWE with the highest contrast (Fig~\ref{fig:eyecatcher}, Right).

In this section we briefly introduce multi-model fitting and discuss how to adapt its components to event data. 
We introduce the notions of spatio-temporal graph of events and model-fitting metric(s) on the graph.

\subsection{Multi-Model Fitting}
\label{sec:multi-model-fitting}

Multi-model fitting is a category of computer vision problems that aim at explaining data using several model instances.
It is a chicken-and-egg problem, often split into two sub-problems: assignment of data to a model instance (i.e., classification or ``labeling'') and model fitting (i.e., parameter estimation).
Examples consist of recognizing and segmenting partially occluded objects in 2D~\cite{winn2006layout} or 3D~\cite{hoiem20073d}, optical
flow estimation~\cite{trobin2008continuous,roth2009discrete,yang2015dense}, and motion segmentation~\cite{isack2012energy}.
These problems aim at assigning a label $l_p$ to each data point $p$, where each label $l \in \mathcal{L}$ corresponds to a model $\mathcal{M}_{l}$ that is consistent with local observations.
The solution to the problem is a labeling configuration $L$ that is locally smooth and globally consistent.
To find such a solution, multi-model fitting problems are naturally formulated as the minimization of an energy $
E = E_{\text{data}} + E_{\text{reg}},$
comprising a data term $E_{\text{data}}$ that measures the inconsistency between the data and the models, 
and a regularizer (smoothness term) $E_{\text{reg}}$ that enforces prior knowledge about the models.
A successful set of solvers consider a graph through the data points 
and seek to partition the graph into the optimal labels \cite{Boykov01iccv,boykov2001fast}.
We also follow this approach.

\subsection{Space-Time Event Graph}
\label{sec:event-graph}
\begin{figure}[t]
  \centering
  \subfigure[t][\small{Events and active pixels.}]{
  \includegraphics[trim={.2cm .5cm .2cm .5cm},clip,width=0.45\columnwidth]{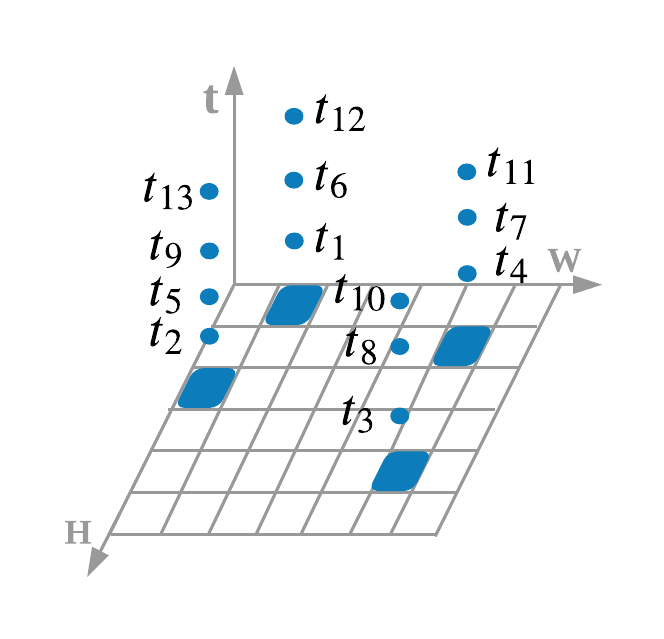}
  \label{fig:active pixels}}
  \subfigure[t][\small{Delaunay triangulation.}]{
  \includegraphics[trim={.2cm .5cm .2cm .5cm},clip,width=0.45\columnwidth]{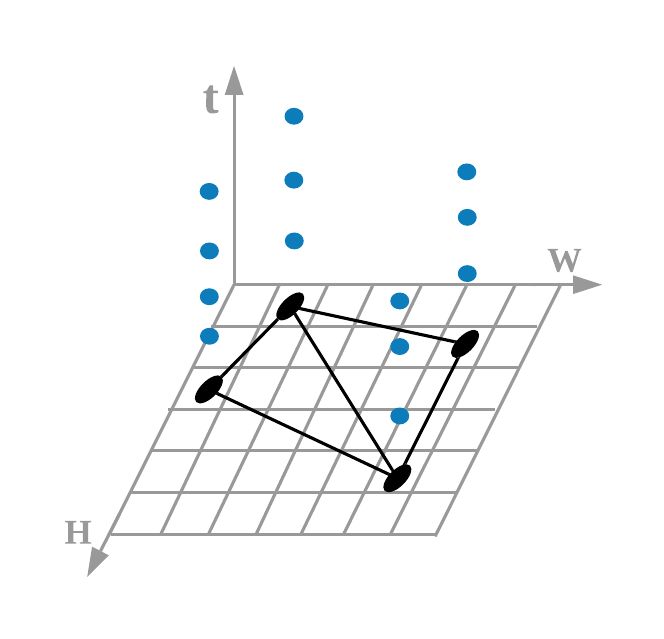}
  \label{fig:delaunay triangulation}}
  \subfigure[t][\small{Connection principle.}]{
  \includegraphics[trim={.2cm .5cm .2cm .5cm},clip,width=0.45\columnwidth]{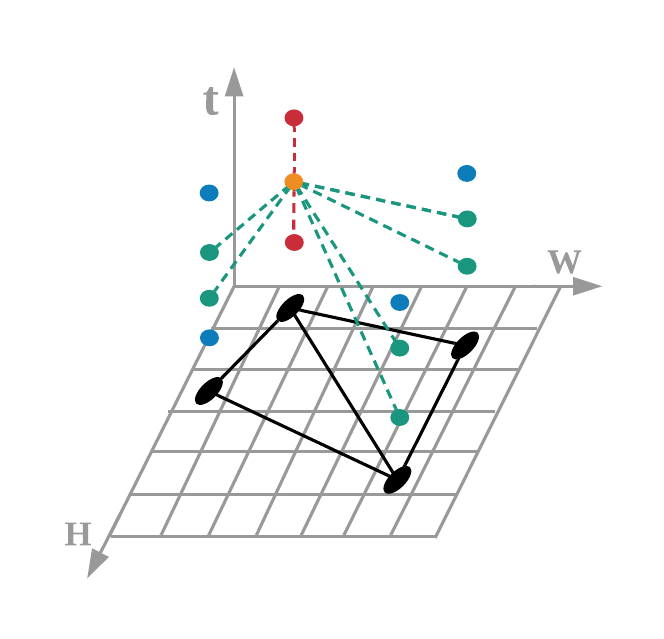}
  \label{fig:connection principle}}
  \subfigure[t][\small{Spatio-temporal graph.}]{
  \includegraphics[trim={.2cm .5cm .2cm .5cm},clip,width=0.45\columnwidth]{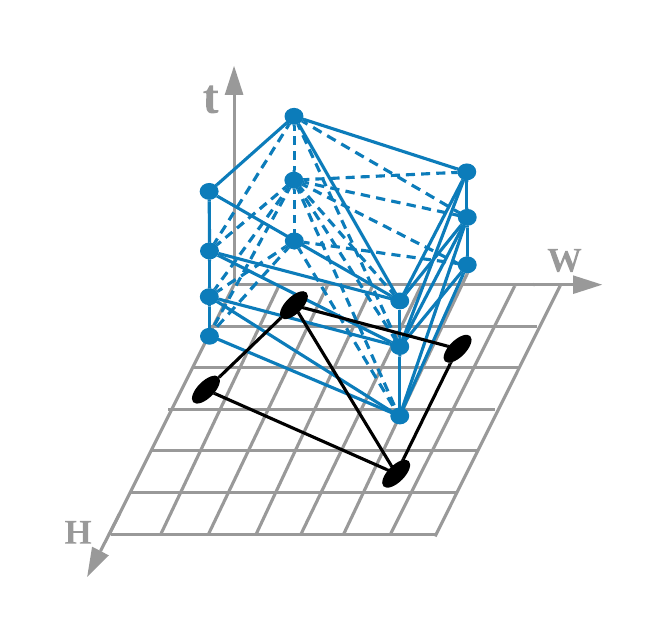}
  \label{fig:resulting ST graph}}
  
  \caption{\emph{Construction of the spatio-temporal graph}.
  (a) Events occur at active pixels (blue squares) and are represented by blue dots along with their corresponding timestamps.
  The indexes indicate the chronological order.
  (b) A Delaunay triangulation (black graph) of the active pixels.
  (c) The connection principle: each event (orange dot) connects to its two temporally-closest events at the same pixel (red dots) and at the neighbouring pixels (green dots) in the Delaunay triangulation.
  (d) The resulting spatio-temporal graph (in blue).}
  \label{fig:spatio-temporal-graph}
\end{figure}

The data points in our problem consist of events $\{e_k\}$ produced by a DVS~\cite{Lichtsteiner08ssc}.
Because events are sparse in the spatio-temporal domain (time-evolving image plane), the underlying graph considered is in general unstructured 
(as opposed to the regular graph of pixel intensities in an image).
To build a spatio-temporal graph for events while keeping a low complexity, we propose to use a Delaunay triangulation~\cite{shewchuk2009general} on the binary image of active events in the space-time volume $\bV$ (Fig.~\ref{fig:spatio-temporal-graph}).

Specifically, given the events in a volume $\mathbf{V}$ of size $W \times H \times \delta t$ (where $W, H$ refer to the width and height of the image plane, and $\delta t$ denotes the time span),
we first compute a binary image %
of event activity (i.e., the pixel is 1 if it contains at least one event, illustrated by blue squares in Fig.~\ref{fig:active pixels}, and it is 0 otherwise).
Then we compute the Delaunay triangulation on the non-zero pixels of this binary image (black dots in %
Fig.~\ref{fig:delaunay triangulation}), which returns a 2D graph (mesh). %
This 2D graph is used to build a space-time (3D) graph for the events 
(see the connection principle in Fig.~\ref{fig:connection principle}).
Each event typically has $2 + 2N$ neighbors in the resulting graph (Fig.~\ref{fig:resulting ST graph}), where $N$ denotes the number of edges that link to the event's pixel location in the binary image (black dots in Fig.~\ref{fig:spatio-temporal-graph}).

There are many possible graphs that can be used to connect the event data points.
The above proposal is inspired in graphs proposed for Markov Random Fields 
built on sets of sparse 2D features \cite{russell2011energy} and is easy to implement from a data structure point of view.

\subsection{Goodness of Fit by Motion Compensation}
\label{sec:review-motion-compensation}
The specification of a metric on the data graph allows us to assess the goodness of fit between the data and the model(s) (\ie, $E_\text{data}$).
Building upon~\cite{Gallego19cvpr}, the goodness of fit is given by the alignment of the events along point trajectories on the image plane,
which constitute a motion model and are parametrized by a parameter vector $\bm$.
Such event alignment is assessed by the strength of the contours of an IWE. %
The contour strength (which is related to image contrast~\cite{gonzalez2004digital}) can be measured with various dispersion metrics, such as variance~\cite{Gallego17ral,Gallego18cvpr,Stoffregen19iccv}, RMS of mean timestamp per pixel~\cite{Zhu19cvpr}
We utilize the variance loss 
as metric for the edges of the event graph.
To make the paper self-contained, we briefly review the idea of motion compensation~\cite{Gallego17ral,Gallego18cvpr}, upon which motion models are fitted.

\textbf{Motion Model Fitting}.
The contrast maximization framework~\cite{Gallego18cvpr} allows us to fit a motion model to a group of events $\cE = \{e_k\}_{k=1}^{N_e}$. 
First, events are geometrically transformed according to a warping function~$\Warp$,
\begin{equation}
\label{eq:warped-events}
e_k \doteq (\bx_k, t_k) \;\mapsto\; e_k^{\prime} \doteq (\bx_k^{\prime}, t_{\text{ref}}),
\end{equation}
leading to a set of warped events $\cE' = \{e_k^{\prime}\}_{k=1}^{N_e}$ at a reference time $t_{\text{ref}}$.
The warping function $\Warp(\bx_k, t_k; \bm) \doteq \bx_k^{\prime}$ implements the image-plane trajectories of a motion model and is parametrized by $\bm$.
Secondly, events $\cE'$ are aggregated into an image (or histogram) of warped events (IWE),
\begin{equation}
\label{eq:IWE}
    I(\bx; \bm) \doteq \sum_{k=1}^{N_e}\delta (\bx - \bx_{k}^{\prime}(\bm)),
\end{equation}
where each pixel $\bx$ counts the number of warped events that fall within it.
In practice, the Dirac function $\delta$ is replaced by a Gaussian $\mathcal{N}(\bx; \mathbf{0},\epsilon^2\text{Id})$ of $\epsilon=1$ pixel width.
Finally, the variance of the IWE 
defines a criterion for model fitting: %
\begin{equation}
\label{eq:argmaxMotionModel}
\bm^\ast = \arg\max_{\bm} \sigma^2(I(\bx; \bm)).
\end{equation}
Like~\cite{Gallego18cvpr}, our formulation supports any type of parametric motion model,
such as \mbox{2-DoF} \emph{(degrees-of-freedom)} flow~\cite{Zhu17icra}, \mbox{3-DoF} rotational motion~\cite{Gallego17ral}, \mbox{4-DoF} model~\cite{Mitrokhin18iros}, etc.

\vspace{0.5ex}
In summary, the above event graph representation and model-fitting criterion 
allow us to formulate the problem of event-based multi-motion segmentation 
as joint estimation over two sets of variables:
\begin{itemize}[noitemsep,nolistsep]
    \item \textbf{Discrete labels}. 
    The segmentation (clustering) is represented by the labeling function $L(e) : \siOmega \times \mathcal{T} \rightarrow \cL = \{1,...,N\}$, 
    which assigns to each event $e$ a label $l \in \cL$ indicating which independently moving object the event belongs to.
    \item \textbf{Motion models}. 
    The motion of the independently moving object is represented by a collection of warps with parameters 
    $\cM = \{\bm_1, ..., \bm_N\}$. 
    Each warp represents the coherent motion of a group of events.
\end{itemize}

\section{Designed Energy for Motion Segmentation}
\label{sec:energy-formualtion}

We propose an energy function that considers jointly the sub-problems of labeling (i.e., segmentation) and model fitting.
The energy function is defined on both unknowns (labeling configuration $L$ and cluster motions $\cM$) as
\begin{equation}
    E(L, \cM) \doteq E_{\text{D}}(L, \cM) + \lambda_{\text{P}}E_{\text{P}}(L) + \lambda_{\text{M}}E_{\text{M}}(L),
\label{eq: overall energy function}
\end{equation}
where $E_{\text{D}}$ denotes the data term and $E_{\text{P}}$, $E_{\text{M}}$ constitute the regularizer. %
Specifically, $E_{\text{P}}$ is the regularizer of the Potts functional \cite{potts1952some} 
and $E_{\text{M}}$ is a label cost term representing the Minimum Descriptor Length (MDL) principle~\cite{delong2012fast}.
The weights $\lambda \geq 0$ balance the contribution of each term.

The energy is designed such that its minimizer achieves the best fit to the data while being spatio-temporally smooth and having the fewest labels (i.e., segments).
We detail the design of each energy term in the upcoming sections.

\subsection{Data Fidelity Term}
\label{subsec: data term}
The data term in~\eqref{eq: overall energy function} is defined on both the discrete labeling variables and the continuous model parameters.
If the labeling variables are fixed, the energy reduces to the data term and the optimal motion model for each cluster can be obtained using the motion compensation scheme (see Section~\ref{subsec: continuous sub problem}).
Hence, we focus on the design of the data term from the perspective of the discrete labeling variables.

To adapt motion compensation~\cite{Gallego18cvpr} to a graph-based energy formulation~\eqref{eq: overall energy function} we need to reformulate it so that:
($i$) energy is minimized instead of maximized, 
($ii$) the fitting cost of a group of events is given by the sum of fitting costs of individual events, i.e., the so-called unary terms~\cite{boykov2001fast}.

The original motion compensation scheme creates an IWE~\eqref{eq:IWE}, on which image contrast is measured.
The IWE is created by warping events according to a certain motion model. %
The higher the IWE contrast, the sharper the IWE, and consequently, higher pixel values (i.e., event accumulation) appear around the edge patterns.
Therefore, the intensity value at each IWE pixel is a proxy for the consistency (goodness of fit) between the model and the events that are warped to that pixel.
To convert the maximization problem into a minimization one and to build the unary costs of the data term, we propose to use the IWE ``negative''~\cite{zhou2020event}.
Once an IWE $I$ is computed~\eqref{eq:IWE}, it is normalized to a fixed range, e.g., $\left[0, 255\right]$, and then its negative is calculated as $\bar{I} = 255 - \hat{I}$, 
where $\hat{\cdot}$ denotes the normalization operation, and $\overline{\cdot}$ the negative operation.
We define the unary term for an event $e_k$ as the value at its warped location, $\bx'_k$ in~\eqref{eq:warped-events}, 
on the IWE negative, namely $\bar{I}(\bx'_k; \bm_l)$.
The data term is defined as the sum of all unary terms:
\begin{equation}
    E_\text{D}(L) \doteq \sum_{l \in \cL}\sum_{e_k \in \mathcal{C}_l} \bar{I}(\bx'_{k}; \bm_l),
\end{equation}
where $\mathcal{C}_l$ denotes the cluster of events with label $l$, 
and $\bar{I}(\cdot\,; \bm_l)$ the IWE negative created using motion model $\bm_l$.

\subsection{Spatially Coherent Labeling}
\label{subsec: potts model term}
As for the regularizer, we simply use a pairwise Potts model term $E_{\text{P}}$ to encourage spatially coherent labeling.
This term is defined only on the discrete labeling variables:
\begin{equation}
    E_{\text{P}}(L) \doteq \sum_{e_i, e_j \in \mathcal{N}} \delta_{L(e_i),L(e_j)},
\end{equation}
where $\mathcal{N}$ denotes the event neighbourhood in the spatio-temporal graph (Section~\ref{sec:event-graph}), 
and $\delta_{m,n}$ is the Kronecker delta (1 if the variables $m,n$ are equal, and 0 otherwise).

\subsection{Encouraging Few Number of Segments}
\label{subsec: MDL term}
To discourage redundancy of the assigned motion models we introduce an MDL term:
\begin{equation}
\label{eq:MDL}
E_{\text{M}}(L) \doteq \sum_{l=1}^{N} \psi(l),
\;\; \psi(l) \doteq \begin{cases} 1 &\mbox{if } \displaystyle{\sum_{e \in \mathcal{C}_l}\delta_{L(e),l} > 0} \\
0 & \mbox{otherwise.}
\end{cases}
\end{equation}
This term penalizes the total number of assigned (active) labels (i.e., segments), 
which encourages it to converge to the actual number of IMOs. %

We describe the optimization procedure for solving the proposed energy in the next section.

\section{Joint Optimization of Motions and Segments}
\label{sec:optimization}
We now discuss how to minimize the proposed energy function \eqref{eq: overall energy function}.
This energy depends on both discrete labeling variables $L$ and continuous motion parameters $\cM$, thus leading to a discrete-continuous optimization problem.
Inspired by efficient solvers used in classical multi-model fitting methods~\cite{delong2012fast,yang2015dense}, 
we employ a block-coordinate descent strategy to optimize $L$ and $\cM$ in an alternating manner.
We present the solution to each sub problem, followed by the initialization strategy.
The overall method is summarized in Algorithm.~\ref{alg: alternating strategy}.

\begin{algorithm}[t]
  \caption{Event-based motion segmentation by discrete-continuous optimization.}
  \label{alg: alternating strategy}
  \begin{algorithmic}[1]
    \State \emph{Input}: $\cE$ events in a space-time volume.
    \State \emph{Output}: event cluster assignments (i.e., labels $L$) and fitted motion per cluster ($\cM$).
    \State Initialize $\cM$, $L$ (Section~\ref{subsec: initialization}).
    \State Create event graph (Section~\ref{sec:event-graph}).
    \State \textbf{Iterate} until convergence:
    \State \quad Fix $\cM$, update $L$ (Section~\ref{subsec: discrete sub problem}, graph cut).
    \State \quad Fix $L$, update $\cM$ (Section~\ref{subsec: continuous sub problem}).
  \end{algorithmic}
\end{algorithm}

\subsection{Segmentation:\! update labels $L$ given motions~$\cM$}
\label{subsec: discrete sub problem}
The overall energy~\eqref{eq: overall energy function} reduces to the sub-problem of discrete labeling when motion models $\cM$ are fixed:
\begin{equation}
    E(L) = E_{\text{D}}(L) + \lambda_{\text{P}}E_{\text{P}}(L) + \lambda_{\text{M}}E_{\text{M}}(L).
\label{eq: discrete sub problem}
\end{equation}
This energy describes a standard %
MRF problem plus an additional MDL term.
The graph-cut method is one of the most widely adopted techniques for solving MRF problems.
The simplest case of graph cut, also known as s-t cut \cite{kolmogorov2004energy}, is typically used for solving a binary classification problem.
A graph is defined connecting the unknowns of the problem, that is, the graph vertices, which must be assigned labels. Two additional vertices called source (s) and sink (t) are defined. 
The minimum s-t cut partitions the vertices of the graph into two disjoint groups (i.e., labels) at the smallest energy cost, which is equivalent to computing the maximum flow from source to sink \cite{ford2015flows}.
The generalization of the minimum s-t-cut problem, such as~\eqref{eq: discrete sub problem}, involves more than two terminals (labels).
For energy functions consisting of a smoothness term with discontinuity-preserving property (e.g., the applied Potts model), the expansion move algorithm \cite{boykov2001fast} can be used as long as the smoothness term is a metric on the space of labels.
The $\alpha$-expansion algorithm loops through the labels $\alpha$ in some order and looks for the lowest energy.
In every iteration, it solves a minimum s-t-cut problem which partitions the vertices into the current-labeling group and the $\alpha$-label group.
An $\alpha$-expansion movement is made according to the s-t cut that leads to the lowest energy.

To minimize \eqref{eq: discrete sub problem}, we apply the $\alpha$-expansion based graph-cut method~\cite{boykov2001fast} combined with the method in~\cite{delong2012fast} to handle the label costs induced by the MDL term.
To accelerate the algorithm, if a motion model is not assigned to any cluster of events, it is removed from the model pool.
The remaining models (label ID) are sorted according to the number of events that belong to the corresponding clusters.

\subsection{Model Fitting:\! update motions $\cM$ given labels~$L$}
\label{subsec: continuous sub problem}
When the label variables are fixed, the energy~\eqref{eq: overall energy function} simplifies to the data term only.
The continuous variables of each motion model can be re-fitted independently from other motion models using the corresponding cluster of events.
The original motion compensation scheme \eqref{eq:argmaxMotionModel} %
is applied for model fitting.

\subsection{Initialization}
\label{subsec: initialization}
Let us show how to initialize the optimization procedure. 
Unlike existing solutions, which either greedily initialize motion models $\cM$~\cite{Stoffregen17acra, Mitrokhin18iros,Stoffregen19iccv} or apply the K-means method on computed event-based optical flow~\cite{Benosman14tnnls} or feature tracks~\cite{parameshwara2020moms},
we propose a simple, direct and effective initialization based on the original motion compensation scheme.

\begin{figure}[t]
\centering
\includegraphics[width=\linewidth]{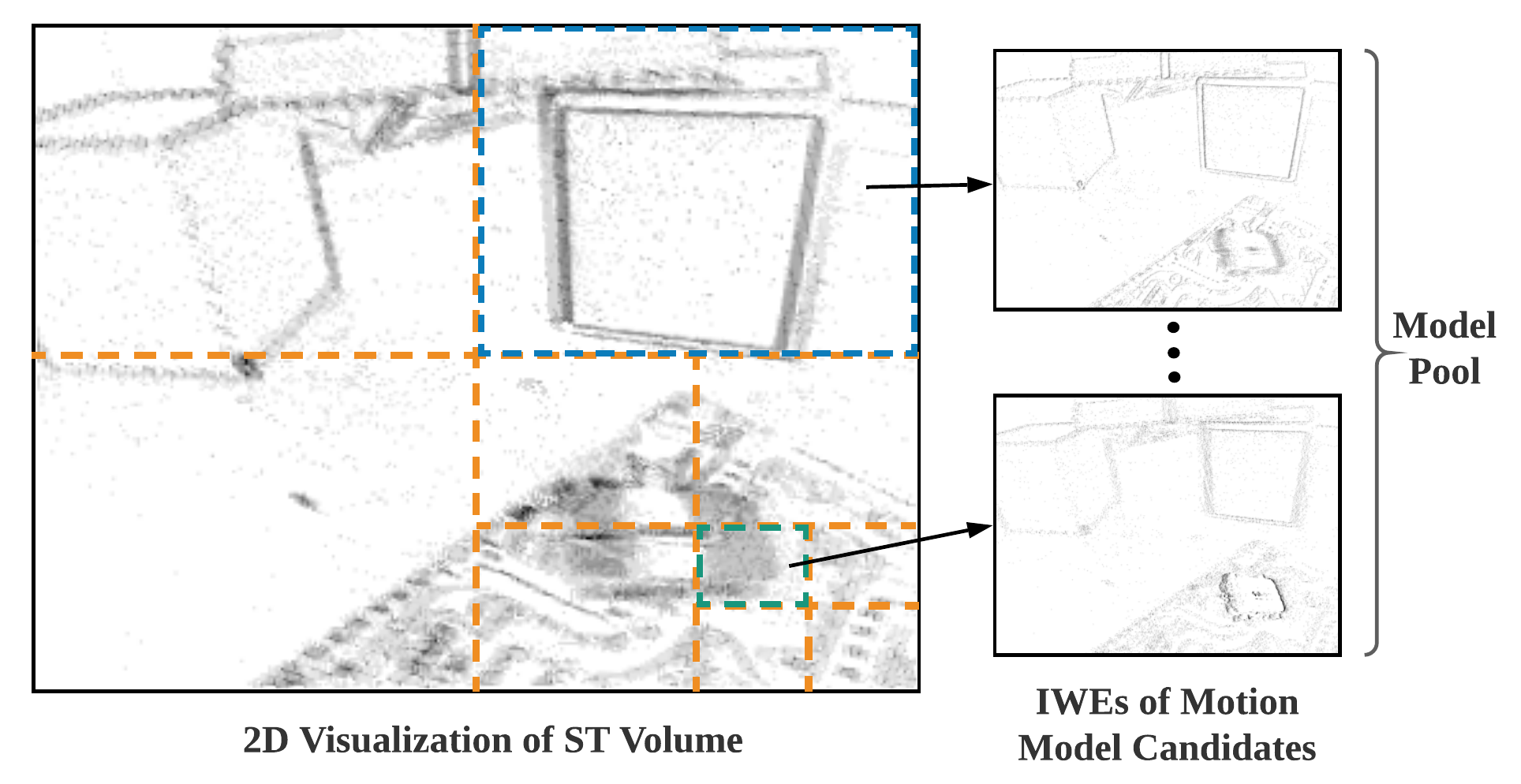}
\caption{Initialization of motion model candidates.
The division operation is illustrated by the orange dashed buckets.
The model pool consists of all IWEs (negative) created using the corresponding model candidates.}
\label{fig: initialization procedure}
\end{figure}

Given a space-time volume $\bV$ of events $\cE$, we carry out an $N$-level subdivision operation.
At level $n \in [0, N-1]$ of the hierarchy, the volume $\bV$ is divided evenly into $4^{n}$ sub-volumes. 
Let us use $N = 4$ as an example, as shown in Fig.~\ref{fig: initialization procedure}, where the blue dashed rectangle illustrates a sub volume at level $n = 1$, while the green one shows a sub-volume at level $n = 3$.
For simplicity, the volume $\bV$ is visualized in 2D, namely by accumulating events on the reference time slice.
After the division operations, we have %
$4^{0} + 4^{1} + \cdots + 4^{3} = 85$ sub-volumes (including the whole volume at the base level of the hierarchy).
By feeding the events in these sub-volumes to the motion compensation scheme, we get a pool of 85 motion model candidates~$\cM$.

This strategy aims at capturing %
IMOs of different size.
As illustrated in Fig.~\ref{fig: initialization procedure}, the blue dashed sub-volume captures one of the boxes in the background, which leads to an IWE with high contrast at the background structures, 
whereas the green dashed sub-volume senses part of a smaller IMO, which leads to an IWE with high contrast around that IMO.
The resulting model pool is used to compute the data term (Section.~\ref{subsec: data term}).

Computational complexity is proportional to the number of event warping operations.
Thus we compare our method against the greedy alternative in terms of the number of warped events.
Assume there are $N_e$ events involved totally and they are induced by the background motion (bg) as well as $m$ IMOs.
Thus, we have $N_e = N_e^{\text{bg}} + \sum_{i=1}^{m} N_e^{\text{IMO}_i}$.
The greedy solution assumes a dominance order of motion models, sorted by the number of events: $N_e^{\text{bg}} \gg N_e^{\text{IMO}_1} \gg ... \gg N_e^{\text{IMO}_m}$.
Under the reasonable assumption of a predominant background motion, $N_e \approx 2\, N_e^{\text{bg}}$, and IMOs accounting for the other half of the events $N_e \approx 2^{i+1}\, N_e^{\text{IMO}_i}$, 
the computational complexity of the greedy method is $O((2 - 2^{-m})\,N_e)$, which is bounded by $O(2\, N_e)$.
The computational complexity of our initialization method depends on the number of subdivision levels $N$, and it is $ O(N\, N_e)$.
While this simplified complexity comparison favors the greedy approach, in practice we found out that our initialization works well by using only the finest level ($n=3$), which has the smallest complexity $O(N_e)$.

\section{Experiments}
\label{sec:evaluation}

In this section we evaluate the proposed algorithm. 
First, we present the datasets and evaluation metrics used (Section~\ref{sec:experim:datasets}).
Second, we provide both quantitative and qualitative results on several datasets %
and compare against state-of-the-art baselines (Section~\ref{sec:experim:quantitative}).
Third, we justify the choice of parameters (Section~\ref{sec:experim:ablation}).
Finally, we analyze the computational efficiency of our implementation (Section~\ref{sec:experim:computational})
and discuss its limitations (Section~\ref{sec:experim:limitations}).

\subsection{Datasets and Evaluation Metrics}
\label{sec:experim:datasets}

We evaluate Algorithm~\ref{alg: alternating strategy} extensively on datasets accompanying recent publications~\cite{Mitrokhin18iros,Stoffregen19iccv,Mitrokhin19iros,almatrafi2020distance}.
We use the same acronym to indicate the method and dataset from a publication, e.g., 
EMSMC denotes the ``Event-based Motion Segmentation by Motion Compensation'' method and its associated dataset, both presented in~\cite{Stoffregen19iccv}. 
The distinction is clear from the context.
\begin{itemize}[noitemsep,nolistsep]
    \item The Extreme Event Dataset (EED) \cite{Mitrokhin18iros} is one of the first open-source datasets used for the research of IMO detection and tracking.
    Besides the camera's ego motion, there are other %
    IMOs (up to three) in each sequence.
    It provides manually-annotated bounding boxes for quantitative evaluation.
    All sequences are collected in a laboratory environment, aiming to demonstrate the outstanding performance of event cameras in HDR scenarios.
    
    \item EVIMO \cite{Mitrokhin19iros} is also collected in a lab environment but with better illumination.
    Up to three IMOs %
    appear in the sequences.
    For evaluation, it provides dense segmentation masks of the IMOs.
    Recently, an extension of the dataset, EVIMO2, has been made available online. 
    
    \item EMSMC \cite{Stoffregen19iccv} provides a set of real-world sequences captured in both indoor and outdoor environments.
    Different from the above two datasets, it consists of a larger number of IMOs and even with non-rigid motions.
    HDR scenarios are also captured in this dataset.
    
    \item DistSurf \cite{almatrafi2020distance} is a dataset collected specifically for evaluating event-based optical flow estimation.
    We find the data is in good quality and can be used for qualitative evaluation of our algorithm.
\end{itemize}
Table~\ref{talble: dataset summary} summarizes the key characteristics of the datasets.

\begin{table}[t]
\caption{Summary of characteristics of the datasets used.}
\label{talble: dataset summary}
\centering
\begin{adjustbox}{max width=\columnwidth}
\setlength{\tabcolsep}{2pt}
\begin{tabular}{lccccc}
\toprule
Dataset & Sensor Type & Sensor Status & \#IMO & Env.  & HDR \\
\midrule
EED~\cite{Mitrokhin18iros}      & DAVIS240    & Moving       & 1-3      & Indoor  & Yes \\
EVIMO~\cite{Mitrokhin19iros}    & DAVIS346    & Moving       & 1-3      & Indoor  & No  \\
EVIMO2~\cite{Mitrokhin19iros}   & Samsung DVS  & Moving       & 1-3  & Indoor  & Yes \\
EMSMC~\cite{Stoffregen19iccv}   & DAVIS240    & Moving       & 1-$N$    & Outdoor & Yes \\
DistSurf~\cite{almatrafi2020distance} & DAVIS346    & Static & 1-$N$    & Indoor  & No \\
\bottomrule %
\end{tabular}
\end{adjustbox}
\end{table}

\vspace{0.5ex}
\textbf{Evaluation Metrics}.
For quantitative evaluation we use two standard metrics.
The first one is \textit{detection rate} based on the overlap between the bounding boxes of detected and labeled objects, which was introduced in~\cite{Mitrokhin18iros} and used ever since. 
It considers the detection result as successful if it meets the following conditions:
\begin{equation}
\label{eq:detectionrate}
\mathcal{B}_D \cap \mathcal{B}_G > 0.5\quad\text{ and }\quad(\mathcal{B}_D \cap \mathcal{B}_G) > (\mathcal{B}_D \cap \overline{\mathcal{B}_G}),
\end{equation}
where $\mathcal{B}_D$ refers to the estimated bounding box (or convex hull), 
$\mathcal{B}_G$ the ground truth bounding box, and $\overline{\cdot}$ denotes the complement of a set.

The second metric is \emph{Intersection over Union} (IoU), which is the most commonly used metric to evaluate the performance of segmentation methods, 
and was proposed for event data in \cite{parameshwara2020moms, Mitrokhin19iros}.
IoU is typically formulated as
\begin{equation}
\label{eq:IoUmetric}
\text{IoU} = (\cS_D \cap \cS_G) / (\cS_D \cup \cS_G),
\end{equation}
where $\cS_D$ refers to the resulting segmentation mask and $\cS_G$ the ground truth mask.
Note that the result of our algorithm consist of sparse warped events with specific labels.
To obtain a dense segmentation mask from a cluster of warped events, we label all pixels in its convex hull identically.

\subsection{Quantitative and Qualitative Evaluation}
\label{sec:experim:quantitative}

\textbf{EED Dataset}.
We first evaluate our algorithm on the EED~\cite{Mitrokhin18iros} dataset using the detection rate metric.
As reported in Table~\ref{tab: EED detection rate}, without having to specify in advance the number of clusters, 
our algorithm outperforms other state-of-the-art solutions~\cite{Mitrokhin18iros,Stoffregen19iccv,parameshwara2020moms} in terms of average detection rate (97.45\%).
The numbers for the baseline methods are taken from the corresponding publications 
in the absence of publicly available source code.
The numbers are close to those in~\cite{Stoffregen19iccv,parameshwara2020moms} partly because the detection rate~\eqref{eq:detectionrate} is a coarse evaluation metric.
Qualitative results are given in Fig.~\ref{fig: qualitative result EED}, where the red rectangles are ground truth bounding boxes.

\begin{table}[t]
\caption{Comparison with state-of-the-art methods on sequences from the \emph{EED dataset}~\cite{Mitrokhin18iros}.
We report detection rate of independently moving objects (in \%), as proposed by~\cite{Mitrokhin18iros}.}
\centering
\begin{adjustbox}{max width=\columnwidth}
\setlength{\tabcolsep}{3pt}
\begin{tabular}{lccccc}
\toprule
             & SOFAS & EED & EMSMC & MOMS-E  & \textbf{Ours} \\
Sequence name & \cite{Stoffregen17acra} & \cite{Mitrokhin18iros} & \cite{Stoffregen19iccv} & \cite{parameshwara2020moms} & (Alg.~\ref{alg: alternating strategy}) \\
\midrule
\emph{Fast drone} & 88.89 & 92.78  & \textbf{96.30}  & - & \textbf{96.30}   \\
\emph{Lighting variation} & 0.00  & 84.52  & 80.51           & - & \textbf{93.51}  \\
\emph{Occlusions}         & 80.00 & 90.83  & 92.31  & - & \textbf{100.00} \\
\emph{What is background?} & 22.08 & 89.21  & \textbf{100.00} & - & \textbf{100.00} \\
\midrule
Average            & 47.74 & 89.34  & 92.28           & 94.20 & \textbf{97.45} \\ 
\bottomrule %
\end{tabular}
\end{adjustbox}
\label{tab: EED detection rate}
\end{table}

\global\long\def\figWidth{0.23\linewidth}
\begin{figure*}[t]
	\centering
    {\small
    \setlength{\tabcolsep}{2pt}
	\begin{tabular}{
	>{\centering\arraybackslash}m{\figWidth} 
	>{\centering\arraybackslash}m{\figWidth}
	>{\centering\arraybackslash}m{\figWidth} 
	>{\centering\arraybackslash}m{\figWidth}}
		\emph{Fast Drone} & \emph{Light Variation} & \emph{Occlusion} & \emph{What is background?}
		\\

		\frame{\includegraphics[width=\linewidth]{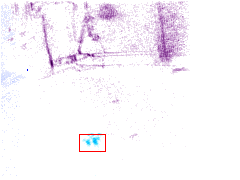}}
		&\frame{\includegraphics[width=\linewidth]{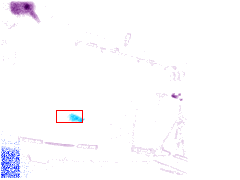}}
		&\frame{\includegraphics[width=\linewidth]{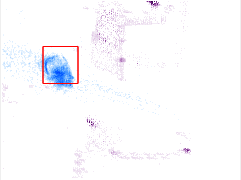}}
		&\frame{\includegraphics[width=\linewidth]{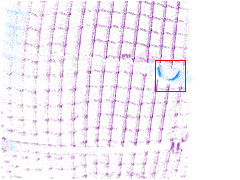}}
		\\
		
		\frame{\includegraphics[width=\linewidth]{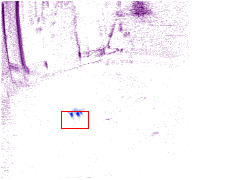}}
		&\frame{\includegraphics[width=\linewidth]{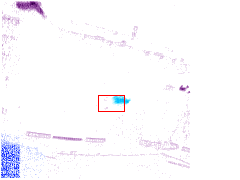}}
		&\frame{\includegraphics[width=\linewidth]{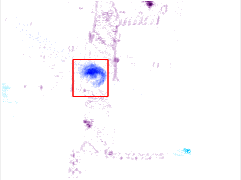}}
		&\frame{\includegraphics[width=\linewidth]{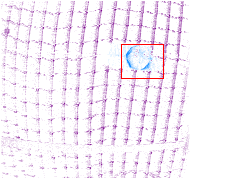}}
		\\
		
		\frame{\includegraphics[width=\linewidth]{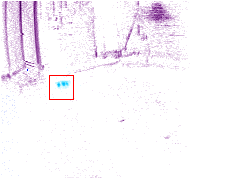}}
		&\frame{\includegraphics[width=\linewidth]{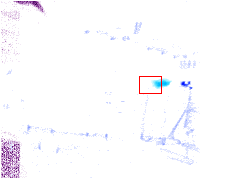}}
		&\frame{\includegraphics[width=\linewidth]{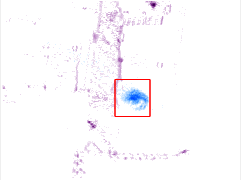}}
		&\frame{\includegraphics[width=\linewidth]{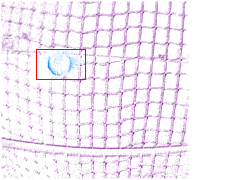}}
		\\
		
		\frame{\includegraphics[width=\linewidth]{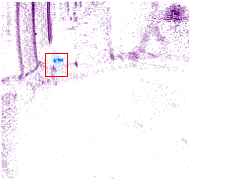}}
		&\frame{\includegraphics[width=\linewidth]{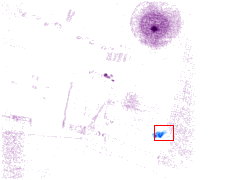}}
		&\frame{\includegraphics[width=\linewidth]{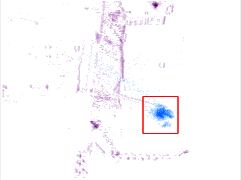}}
		&\frame{\includegraphics[width=\linewidth]{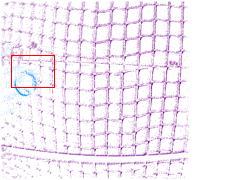}}
		\\

	\end{tabular}
	}
	\caption{Segmentation results on EED dataset~\cite{Mitrokhin18iros}. 
	Time runs from top to bottom.
	The ground truth bounding boxes (in red) show the 2D location of the IMOs. 
	Note that such boxes are manually annotated on the DAVIS~\cite{Brandli14ssc} grayscale images, which are not perfectly time-aligned with the events. 
	Hence, offsets are witnessed for fast-moving IMOs.}
	\label{fig: qualitative result EED}
\end{figure*}

\textbf{EVIMO Dataset}.
A second quantitative evaluation is performed on the EVIMO dataset using the IoU metric~\eqref{eq:IoUmetric}.
As shown in Table~\ref{tab: IoU evaluation}, our algorithm performs the second best (only 0.19\% lower than the best one) among all four segmentation methods,
and about 2\% better than the next best method.
The numbers for the baseline methods are taken from~\cite{parameshwara2020moms}.
Note that Tables~\ref{tab: EED detection rate} and~\ref{tab: IoU evaluation} report results using two different metrics and datasets, so they are not directly related.
During the experiments we found several issues that may deteriorate the IoU score.
First, the ground truth masks are not perfectly aligned with objects in the raw images 
because of inaccurate CAD models and IMO poses produced by the motion capture system.
Second, our labeled IWE and segmentation masks are computed using undistorted events (using camera calibration); then the masks are warped to the raw (distorted) image plane where the IoU score is calculated.
Undistortion leads to information loss near image boundaries. %
Third, some of the IMOs are hanged by the person who was holding the event camera, thus, the IMOs sometimes undergo the same motion as the camera.
In such a case, IMOs are labeled as background motion and no IoU score is calculated.
\global\long\def\figWidth{0.153\linewidth}
\begin{figure*}[t]
	\centering
    {\small
    \setlength{\tabcolsep}{2pt}
	\begin{tabular}{
	>{\centering\arraybackslash}m{0.4cm}
	>{\centering\arraybackslash}m{\figWidth} 
	>{\centering\arraybackslash}m{\figWidth}
	>{\centering\arraybackslash}m{\figWidth}
	>{\centering\arraybackslash}m{\figWidth}
	>{\centering\arraybackslash}m{\figWidth}
	>{\centering\arraybackslash}m{\figWidth}}
		\\

		\rotatebox{90}{\makecell{Ours: IWE}}
		&\frame{\includegraphics[width=\linewidth]{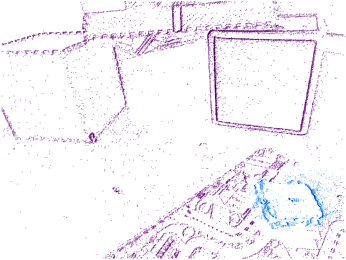}}
		&\frame{\includegraphics[width=\linewidth]{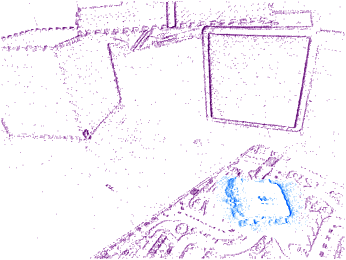}}
		&\frame{\includegraphics[width=\linewidth]{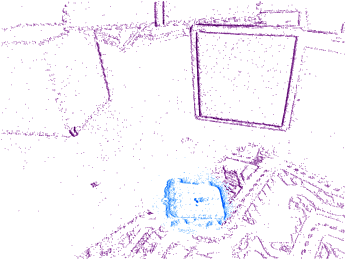}}
		&\frame{\includegraphics[width=\linewidth]{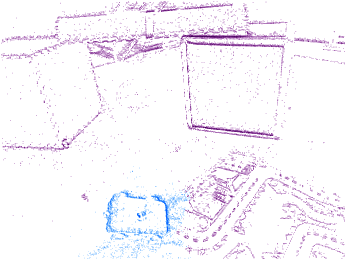}}
		&\frame{\includegraphics[width=\linewidth]{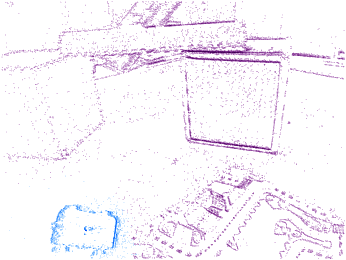}}
		&\frame{\includegraphics[width=\linewidth]{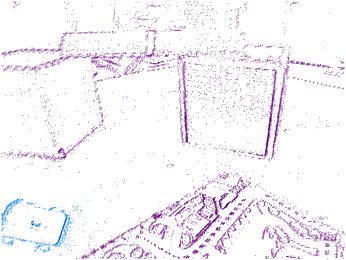}}
		\\
		
		\rotatebox{90}{\makecell{Ours:  masks}}
		&\frame{\includegraphics[width=\linewidth]{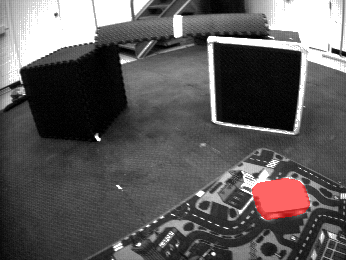}}
		&\frame{\includegraphics[width=\linewidth]{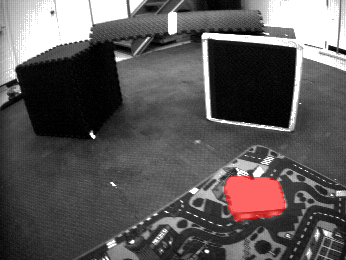}}
		&\frame{\includegraphics[width=\linewidth]{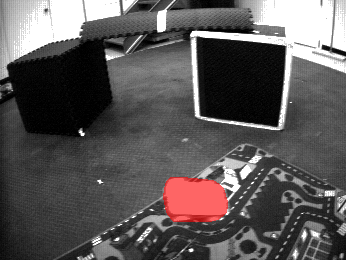}}
		&\frame{\includegraphics[width=\linewidth]{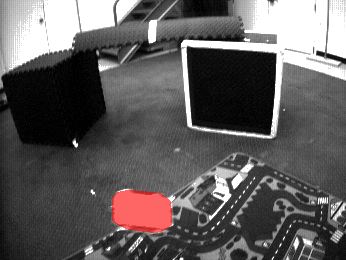}}
		&\frame{\includegraphics[width=\linewidth]{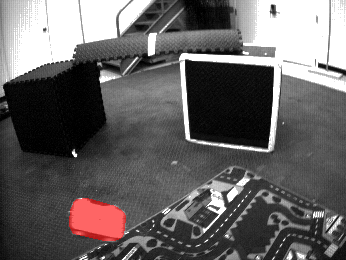}}
		&\frame{\includegraphics[width=\linewidth]{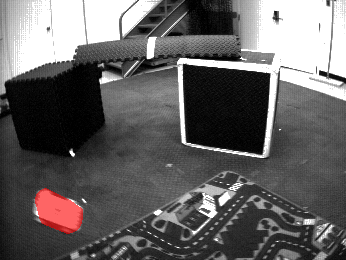}}
		\\

		\rotatebox{90}{\makecell{\cite{parameshwara2020moms}} IWE}
		&\frame{\includegraphics[width=\linewidth]{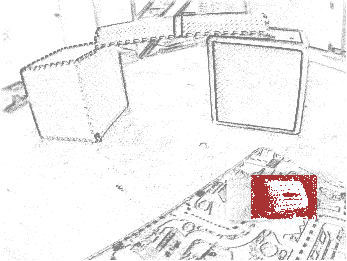}}
		&\frame{\includegraphics[width=\linewidth]{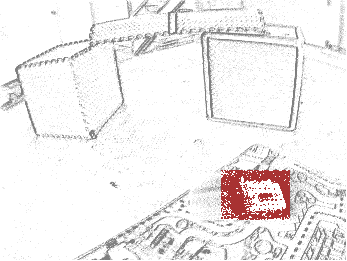}}
		&\frame{\includegraphics[width=\linewidth]{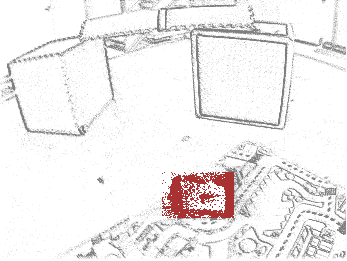}}
		&\frame{\includegraphics[width=\linewidth]{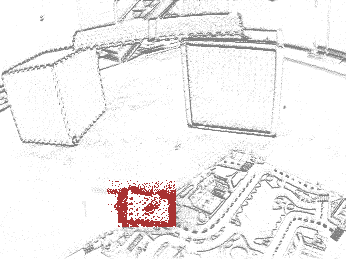}}
		&\frame{\includegraphics[width=\linewidth]{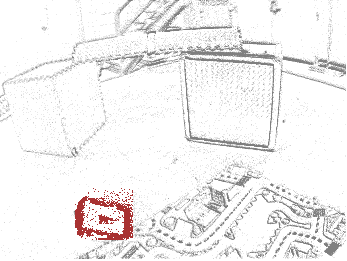}}
		&\frame{\includegraphics[width=\linewidth]{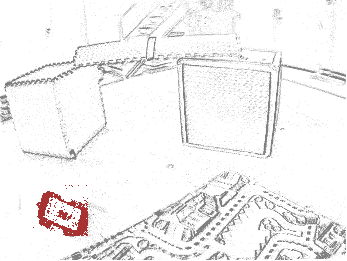}}
		\\\addlinespace[1ex]
		
		\rotatebox{90}{\makecell{Ours: IWE}}
		&\frame{\includegraphics[width=\linewidth]{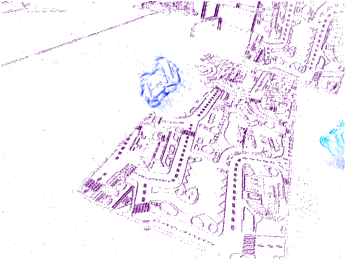}}
		&\frame{\includegraphics[width=\linewidth]{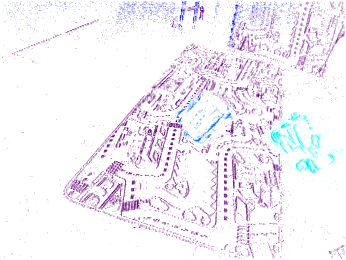}}
		&\frame{\includegraphics[width=\linewidth]{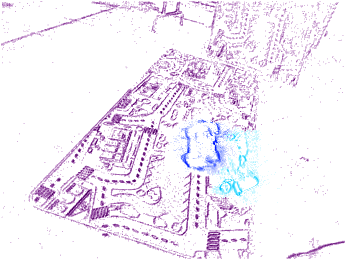}}
		&\frame{\includegraphics[width=\linewidth]{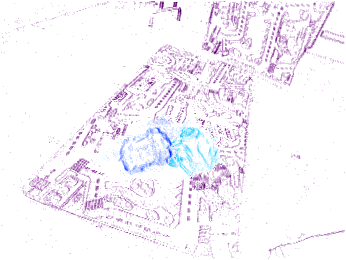}}
		&\frame{\includegraphics[width=\linewidth]{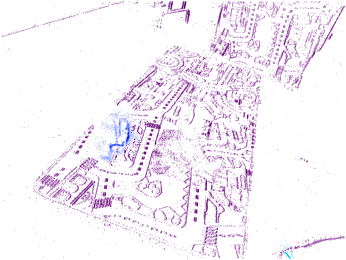}}
		&\frame{\includegraphics[width=\linewidth]{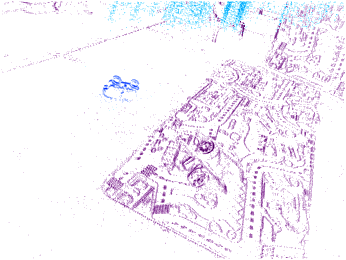}}
		\\
		
		\rotatebox{90}{\makecell{Ours:  masks}}
		&\frame{\includegraphics[width=\linewidth]{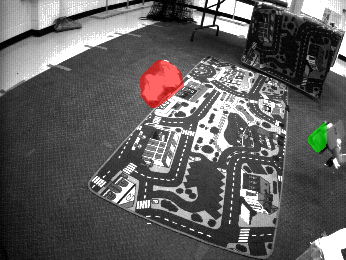}}
		&\frame{\includegraphics[width=\linewidth]{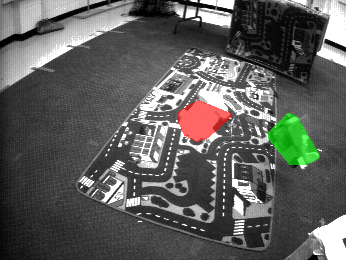}}
		&\frame{\includegraphics[width=\linewidth]{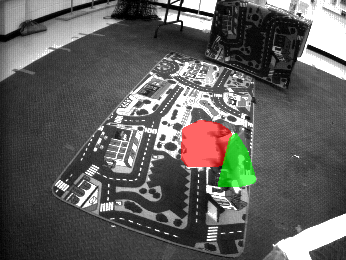}}
		&\frame{\includegraphics[width=\linewidth]{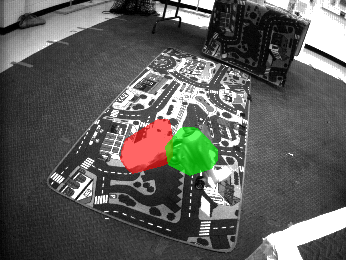}}
		&\frame{\includegraphics[width=\linewidth]{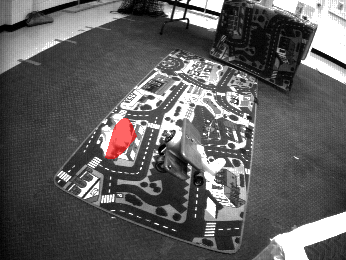}}
		&\frame{\includegraphics[width=\linewidth]{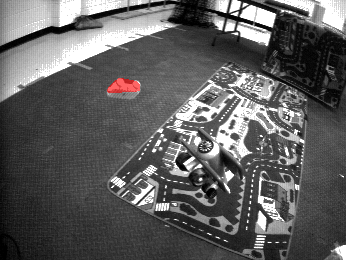}}
		\\
	
		\rotatebox{90}{\makecell{\cite{parameshwara2020moms}} IWE}
		&\frame{\includegraphics[width=\linewidth]{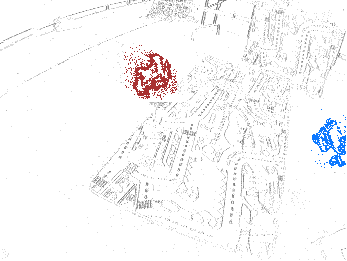}}
		&\frame{\includegraphics[width=\linewidth]{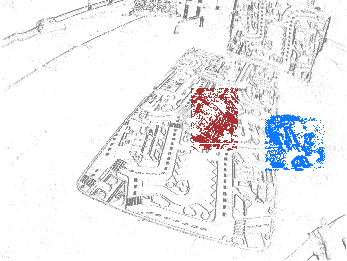}}
		&\frame{\includegraphics[width=\linewidth]{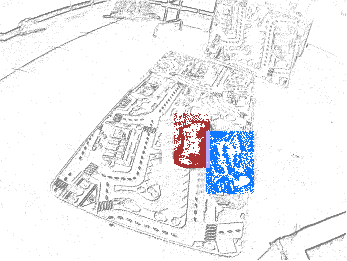}}
		&\frame{\includegraphics[width=\linewidth]{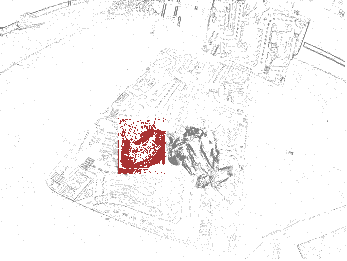}}
		&\frame{\includegraphics[width=\linewidth]{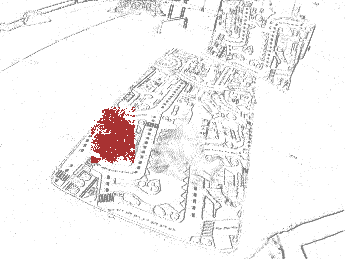}}
		&\frame{\includegraphics[width=\linewidth]{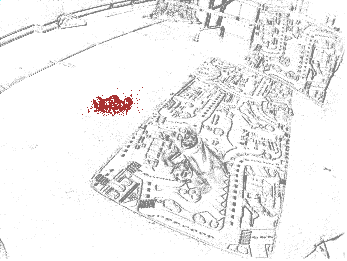}}
		\\
		
	\end{tabular}
	}
	\caption{Segmentation results on the EVIMO dataset~\cite{Mitrokhin19iros}, 
	on sequences \emph{Boxes} (rows 1--3) and \emph{Table} (rows 4--6).
	Time runs from left to right. 
	Our labeled IWEs (rows 1 \& 4) are displayed in undistorted coordinates (known camera calibration) 
	while the dense segmentation masks (rows 2 \& 5) are shown on the raw (distorted) grayscale images from the DAVIS~\cite{Brandli14ssc}.
	Grayscale images are used for visualization only.
	The labeled IWEs from \cite{parameshwara2020moms} (rows 3 \& 6) are shown also on raw coordinates, as provided in~\cite{parameshwara2020moms}.
	Their color convention is: gray for the background (ego-motion) and dark red or blue for the moving objects.
	The corresponding colors in our IWEs are magenta, blue and cyan, respectively.}
	\label{fig: qualtitative evaluation EVIMO}
\end{figure*}
\begin{table}[t]
\caption{Comparison with state-of-the-art methods on the EVIMO dataset~\cite{Mitrokhin19iros}.
We report the IoU metric (in \%), as given by \cite{parameshwara2020moms, Mitrokhin19iros}.}
\centering
\begin{adjustbox}{max width=\columnwidth}
\setlength{\tabcolsep}{4pt}
\begin{tabular}{c|cccl}
\toprule
& EVIMO & EVDodgeNet & MOMS-E & \textbf{Ours} \\
& method \cite{Mitrokhin19iros} & \cite{Sanket20icra} & \cite{parameshwara2020moms} &  \\
\midrule
\textit{EVIMO dataset} &  \textbf{77.00}     & 65.76      & 74.82  &  76.81    \\
\toprule
\end{tabular}
\end{adjustbox}
\label{tab: IoU evaluation}
\vspace{-1ex}
\end{table}
\global\long\def\figWidth{0.155\linewidth}
\begin{figure*}[h!]
	\centering
    {\small
    \setlength{\tabcolsep}{2pt}
	\begin{tabular}{
	>{\centering\arraybackslash}m{\figWidth} 
	>{\centering\arraybackslash}m{\figWidth}
	>{\centering\arraybackslash}m{\figWidth}
	>{\centering\arraybackslash}m{\figWidth} 
	>{\centering\arraybackslash}m{\figWidth}
	>{\centering\arraybackslash}m{\figWidth}}
		\multicolumn{2}{c}{\emph{Traffic scene}} &
		\multicolumn{2}{c}{\emph{Facing the Sun}} &
		\multicolumn{2}{c}{\emph{Building \& car}}
		\\\cmidrule(l{6mm}r{6mm}){1-2} \cmidrule(l{6mm}r{6mm}){3-4} \cmidrule(l{6mm}r{6mm}){5-6} %
		
		EMSMC~\cite{Stoffregen19iccv} & 
		Ours (Alg.~\ref{alg: alternating strategy})&
		EMSMC~\cite{Stoffregen19iccv} &
		Ours (Alg.~\ref{alg: alternating strategy})&
		EMSMC~\cite{Stoffregen19iccv} &
		Ours (Alg.~\ref{alg: alternating strategy})
		\\[0.3ex]%

		\frame{\includegraphics[trim={0.6cm 0.1cm 0.25cm 0.5cm},clip,width=\linewidth]{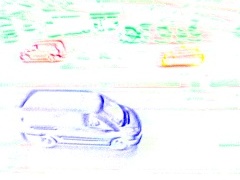}}
		&\frame{\includegraphics[width=\linewidth]{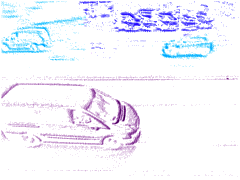}}
		&\frame{\includegraphics[trim={0.6cm 0.1cm 0.25cm 0.5cm},clip,width=\linewidth]{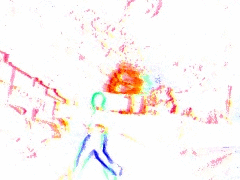}}
		&\frame{\includegraphics[width=\linewidth]{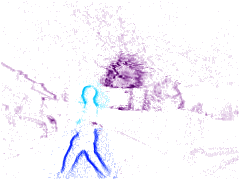}}
		&\frame{\includegraphics[trim={0.65cm 0.1cm 0.25cm 0.5cm},clip,width=\linewidth]{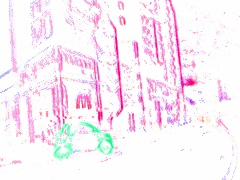}}
		&\frame{\includegraphics[width=\linewidth]{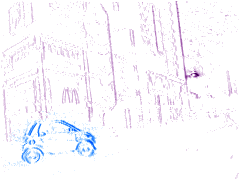}}
		\\
		
		\frame{\includegraphics[trim={0.6cm 0.1cm 0.25cm 0.5cm},clip,width=\linewidth]{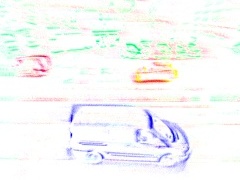}}
		&\frame{\includegraphics[width=\linewidth]{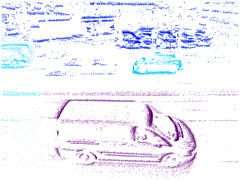}}
		&\frame{\includegraphics[trim={0.6cm 0.1cm 0.25cm 0.5cm},clip,width=\linewidth]{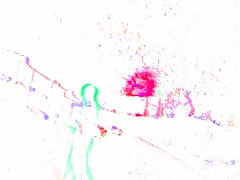}}
		&\frame{\includegraphics[width=\linewidth]{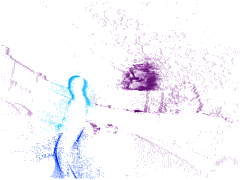}}
		&\frame{\includegraphics[trim={0.65cm 0.1cm 0.25cm 0.5cm},clip,width=\linewidth]{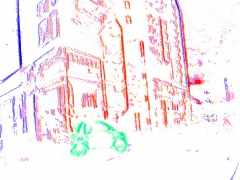}}
		&\frame{\includegraphics[width=\linewidth]{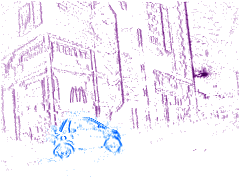}}
		\\
		
		\frame{\includegraphics[trim={0.6cm 0.1cm 0.25cm 0.5cm},clip,width=\linewidth]{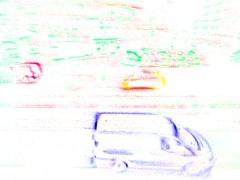}}
		&\frame{\includegraphics[width=\linewidth]{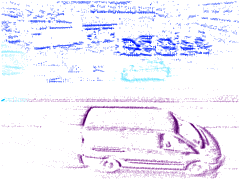}}
		&\frame{\includegraphics[trim={0.6cm 0.1cm 0.25cm 0.5cm},clip,width=\linewidth]{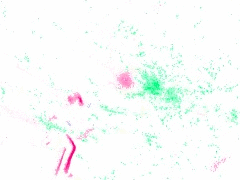}}
		&\frame{\includegraphics[width=\linewidth]{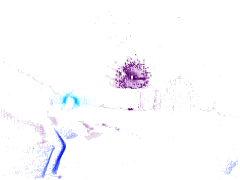}}
		&\frame{\includegraphics[trim={0.65cm 0.1cm 0.25cm 0.5cm},clip,width=\linewidth]{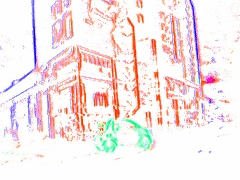}}
		&\frame{\includegraphics[width=\linewidth]{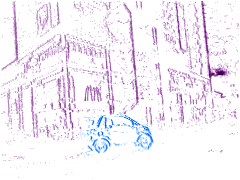}}
		\\

		\frame{\includegraphics[trim={0.6cm 0.1cm 0.25cm 0.5cm},clip,width=\linewidth]{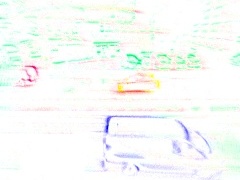}}
		&\frame{\includegraphics[width=\linewidth]{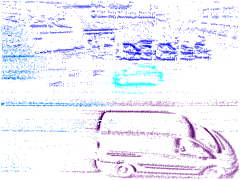}}
		&\frame{\includegraphics[trim={0.6cm 0.1cm 0.25cm 0.5cm},clip,width=\linewidth]{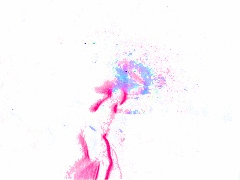}}
		&\frame{\includegraphics[width=\linewidth]{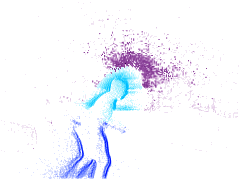}}
		&\frame{\includegraphics[trim={0.65cm 0.1cm 0.25cm 0.5cm},clip,width=\linewidth]{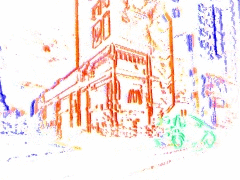}}
		&\frame{\includegraphics[width=\linewidth]{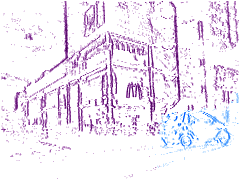}}
		\\
	\end{tabular}
	}
	\caption{Qualitative comparison with EMSMC~\cite{Stoffregen19iccv} on sequences from that reference. 
	Time runs from top to bottom.
	Images in odd columns are courtesy of~\cite{Stoffregen19iccv}, with clusters represented in various colors due to the over-segmentation in EMSMC.
	\label{fig:emsmc:compare}}
\end{figure*}

\global\long\def\figWidth{0.155\linewidth}
\begin{figure*}[t]
	\centering
    {\small
    \setlength{\tabcolsep}{2pt}
	\begin{tabular}{
	>{\centering\arraybackslash}m{\figWidth} 
	>{\centering\arraybackslash}m{\figWidth}
	>{\centering\arraybackslash}m{\figWidth}
	>{\centering\arraybackslash}m{\figWidth} 
	>{\centering\arraybackslash}m{\figWidth}
	>{\centering\arraybackslash}m{\figWidth}}
		\emph{Fan \& coin} & \emph{Cars} & \emph{Hands} & \emph{Toppling Box} & \emph{Slope} & \emph{Cast}
		\\[0.3ex]%

		\frame{\includegraphics[width=\linewidth]{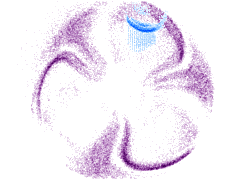}}
		&\frame{\includegraphics[width=\linewidth]{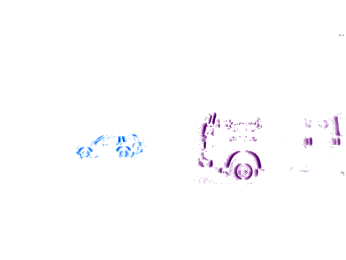}}
		&\frame{\includegraphics[width=\linewidth]{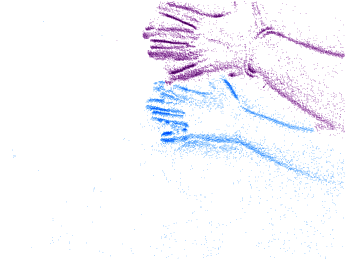}}
		&\frame{\includegraphics[width=\linewidth]{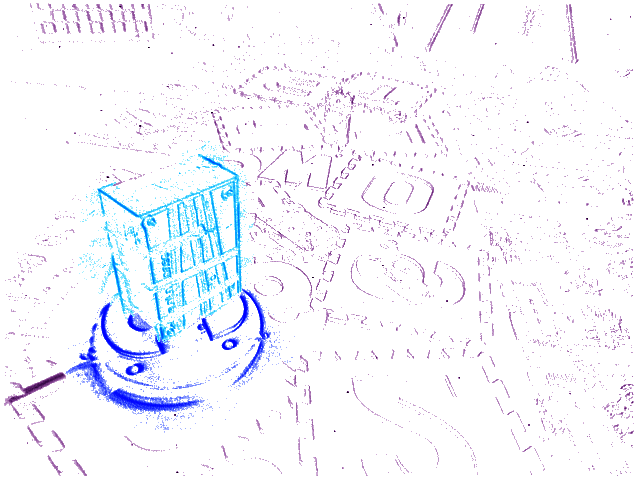}}
		&\frame{\includegraphics[width=\linewidth]{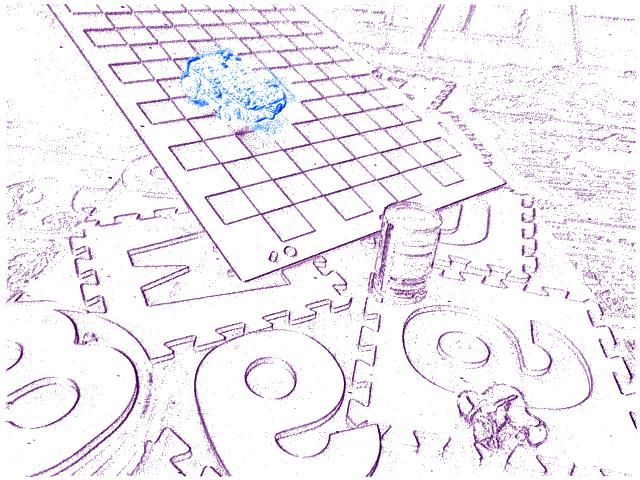}}
		&\frame{\includegraphics[width=\linewidth]{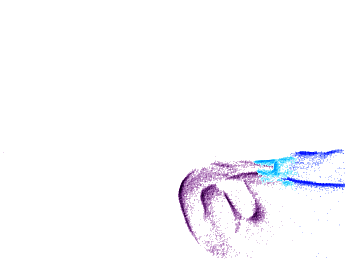}}
		\\
		
		\frame{\includegraphics[width=\linewidth]{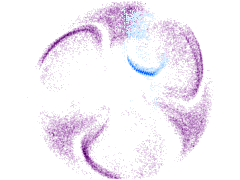}}
		&\frame{\includegraphics[width=\linewidth]{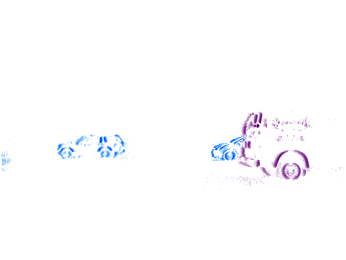}}
		&\frame{\includegraphics[width=\linewidth]{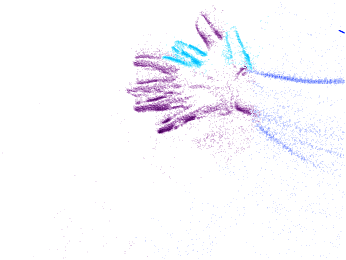}}
		&\frame{\includegraphics[width=\linewidth]{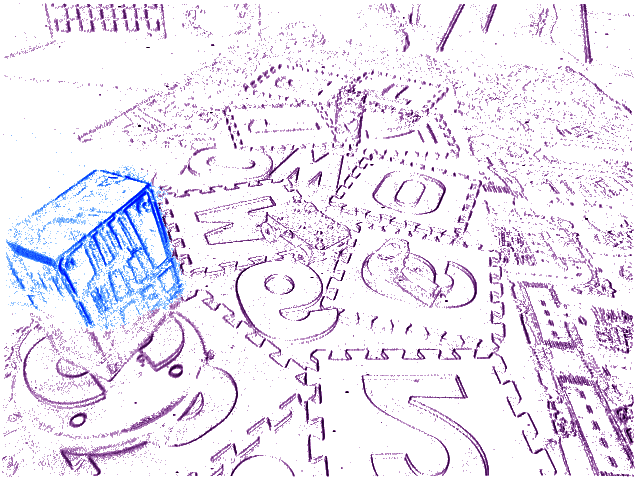}}
		&\frame{\includegraphics[width=\linewidth]{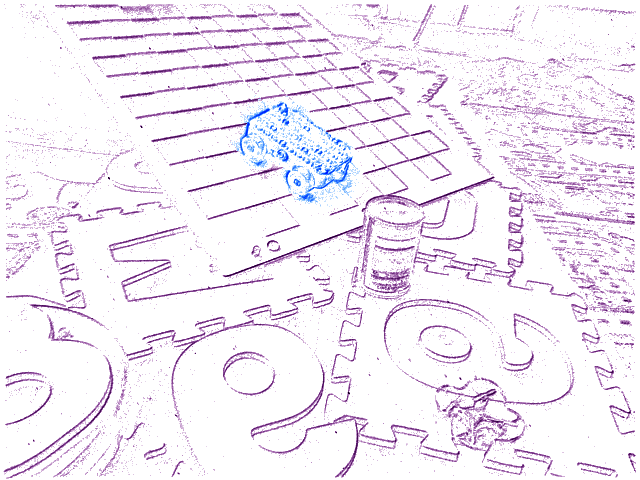}}
		&\frame{\includegraphics[width=\linewidth]{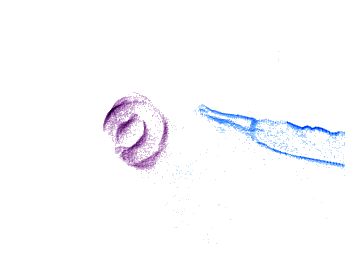}}
		\\
		
		\frame{\includegraphics[width=\linewidth]{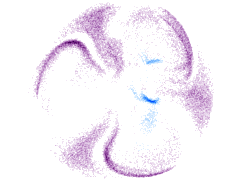}}
		&\frame{\includegraphics[width=\linewidth]{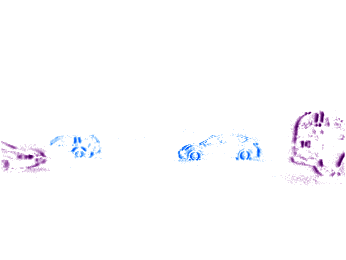}}
		&\frame{\includegraphics[width=\linewidth]{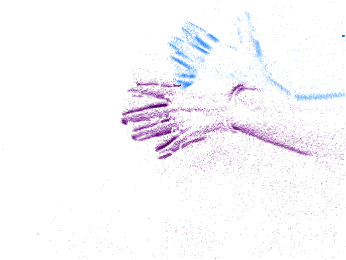}}
		&\frame{\includegraphics[width=\linewidth]{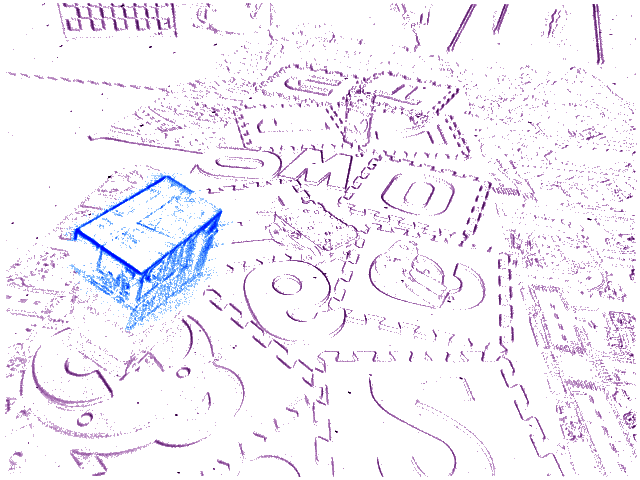}}
		&\frame{\includegraphics[width=\linewidth]{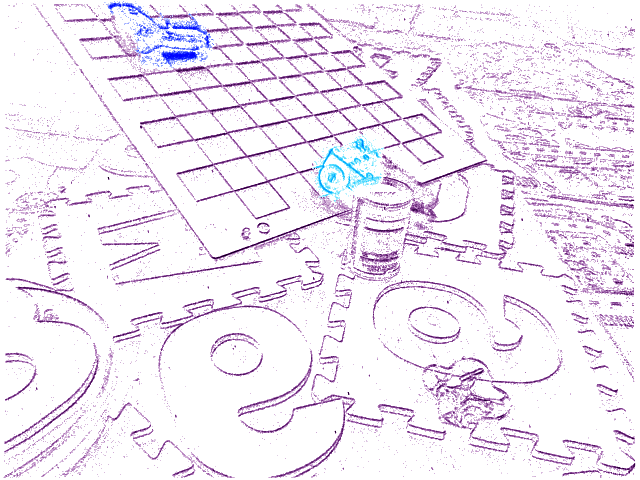}}
		&\frame{\includegraphics[width=\linewidth]{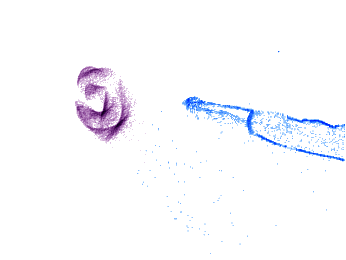}}
		\\
		
		\frame{\includegraphics[width=\linewidth]{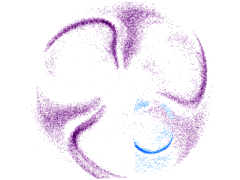}}
		&\frame{\includegraphics[width=\linewidth]{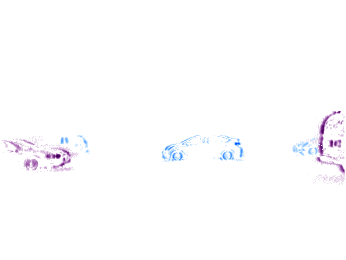}}
		&\frame{\includegraphics[width=\linewidth]{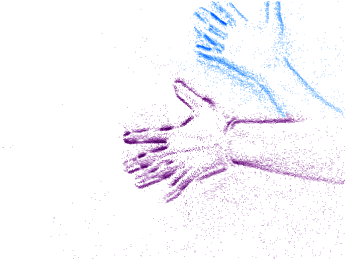}}
		&\frame{\includegraphics[width=\linewidth]{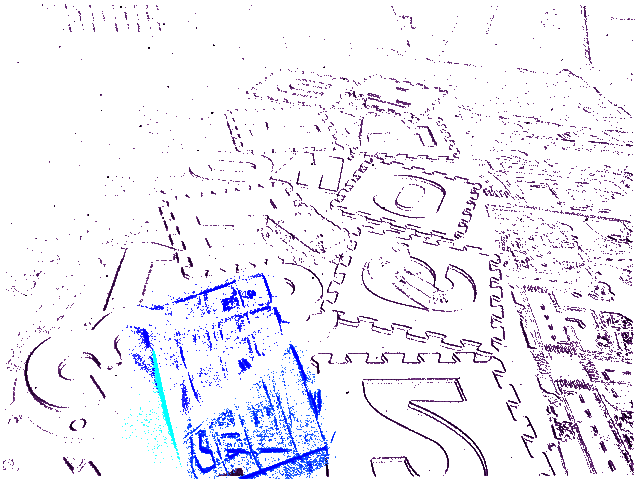}}
		&\frame{\includegraphics[width=\linewidth]{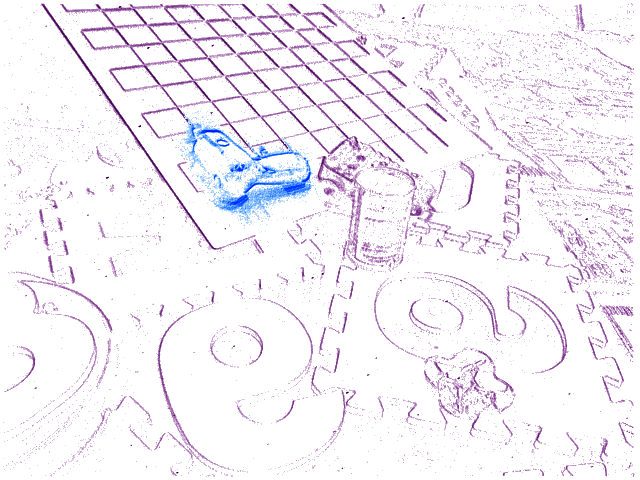}}
		&\frame{\includegraphics[width=\linewidth]{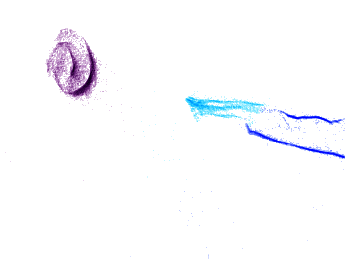}}
		\\
	\end{tabular}
	}
	\caption{Results of our method on sequences from EMSMC~\cite{Stoffregen19iccv} (column 1), DistSurf~\cite{almatrafi2020distance} (columns 2-3), EVIMO2~\cite{Mitrokhin19iros} (columns 4-5), and our data (column 6).
	Time runs from top to bottom.
	Our segmentation method is able to identify IMOs in a variety of scenes (from different datasets, with stationary or moving cameras) and with various sensor resolutions (examples with event cameras from $240\times 180$ pixels to $640\times 480$ pixels).
	}
	\label{fig:real:more}
\end{figure*}

Exemplary results are given in Fig.~\ref{fig: qualtitative evaluation EVIMO}, where both labeled IWEs and their corresponding dense segmentation masks are visualized.
In the \emph{boxes} sequence, the toy car traverses from right to left and can be continuously detected as a moving object.
There are two IMOs in the \emph{table} sequence (toy car and plane).
The two move against each other and meet in the middle.
They are successfully detected even when they are partially overlapped.
The toy plane stays almost still at the end of the sequence, which explains why it is labeled as background.
The IWE around the area of the plane looks as sharp as the background, which also indicates that the plane stays still.
The figure also includes qualitative comparisons against~\cite{parameshwara2020moms} in terms of the labeled IWEs. 
Despite the different coordinates used, our results look overall sharper and do not exhibit the rectangular segmentation boundaries present in~\cite{parameshwara2020moms} (caused by naively labeling events according to whether or not they are inside the convex hull of the cluster features). 
In the middle of the \emph{table} sequence, our method still detects the toy plane as an IMO (before it slows down and merges with the background), whereas \cite{parameshwara2020moms} does not.

In addition, the authors of \cite{Mitrokhin19iros} just released a new dataset, EVIMO2, using event-based cameras of VGA resolution ($640\times 480$ pixels).
The dataset can be used for evaluation of event-based object segmentation, motion segmentation, and structure from motion.
The multi-camera system consists of a Samsung Gen3 DVS~\cite{Son17isscc}, 
two Prophesee CD Gen3 event cameras~\cite{propheseeevk} and a Flea3 RGB camera.
We test our method on the events from the Samsung DVS and provide qualitative results in Fig.~\ref{fig:real:more}.
We also compute the IoU score (64.38 \%), and the main reason for the less accurate segmentation than in EVIMO is the motion patterns: 
the objects in EVIMO2 undergo more frequent 3D rotations than in EVIMO.
Consequently, the 2D appearance of the objects on the image plane continuously changes, causing self-occlusions that are difficult to model with current image-plane motion models. 
To obtain best results in this scenario, knowledge of the 3D shape and appearance of the IMO would be required, which is left as future work.
In spite of this, the qualitative results in Fig.~\ref{fig:real:more} (columns 4 and~5) show good segmentation results.

\global\long\def\figWidth{0.155\linewidth}
\begin{figure*}[t]
	\centering
    {\small
    \setlength{\tabcolsep}{2pt}
	\begin{tabular}{
	>{\centering\arraybackslash}m{0.3cm}
	>{\centering\arraybackslash}m{\figWidth} 
	>{\centering\arraybackslash}m{\figWidth}
	>{\centering\arraybackslash}m{\figWidth}
	>{\centering\arraybackslash}m{\figWidth} 
	>{\centering\arraybackslash}m{\figWidth}
	>{\centering\arraybackslash}m{\figWidth}}
		\emph{} &
		\emph{$\lambda_{\text{M}} = 1000$} & 
		\emph{$\lambda_{\text{M}} = 2000$} &
		\emph{$\lambda_{\text{M}} = 4000$} &
		\emph{$\boldsymbol{\lambda_\text{M} = 8000}$} &
		\emph{$\boldsymbol{\lambda_\text{M} = 16000}$} &
		\emph{$\lambda_{\text{M}} = 32000$}
		\\[1.3ex]%
        
        \rotatebox{90}{\makecell{\emph{Corridor}}}&
		\frame{\includegraphics[width=\linewidth]{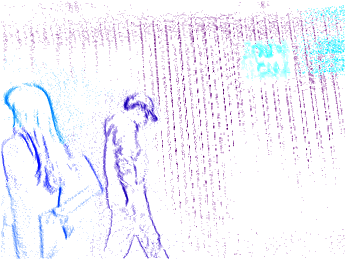}}&
		\frame{\includegraphics[width=\linewidth]{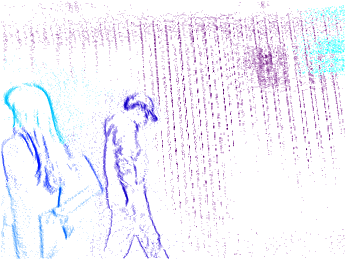}}&
		\frame{\includegraphics[width=\linewidth]{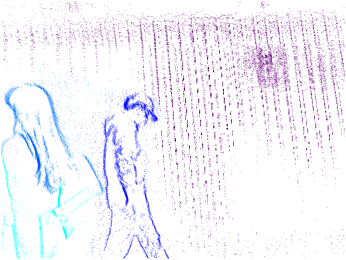}}&
		\frame{\includegraphics[width=\linewidth]{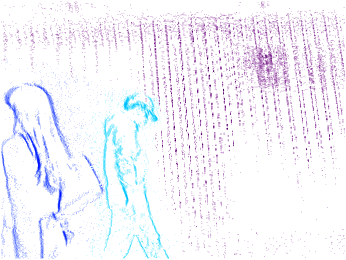}}&
		\frame{\includegraphics[width=\linewidth]{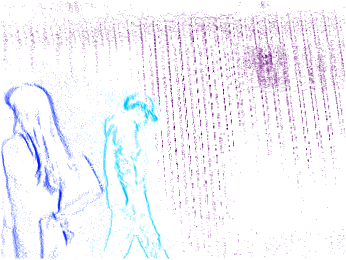}}&
		\frame{\includegraphics[width=\linewidth]{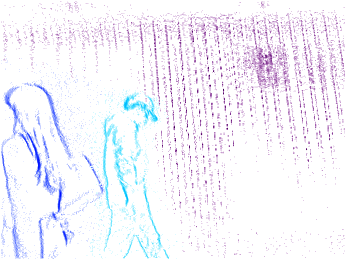}}
		\\
		
		&
		\emph{7 models} & 
		\emph{6 models} & 
		\emph{4 models} & 
		\emph{3 models} & 
		\emph{3 models} & 
		\emph{3 models} 
		\\[1.5ex]%

		\rotatebox{90}{\makecell{\emph{Table}}}&
		\frame{\includegraphics[width=\linewidth]{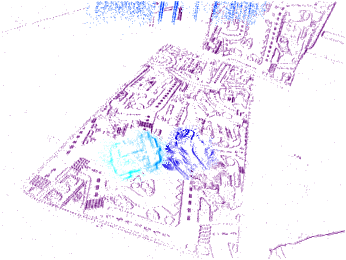}}&
		\frame{\includegraphics[width=\linewidth]{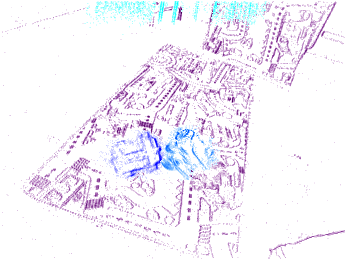}}&
		\frame{\includegraphics[width=\linewidth]{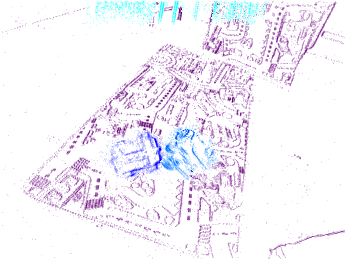}}&
		\frame{\includegraphics[width=\linewidth]{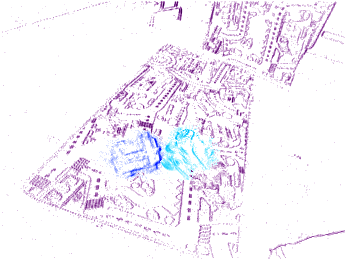}}&
		\frame{\includegraphics[width=\linewidth]{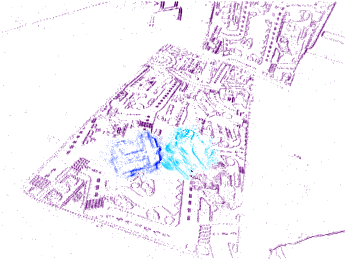}}&
		\frame{\includegraphics[width=\linewidth]{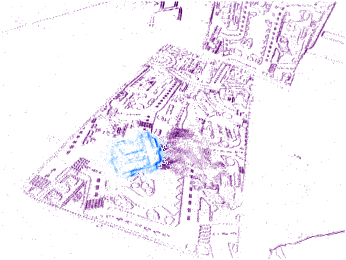}}
		\\
		
	     &
		\emph{5 models} & 
		\emph{4 models} & 
		\emph{4 models} & 
		\emph{3 models} & 
		\emph{3 models} & 
		\emph{2 models} 
		\\[0.3ex]%
	\end{tabular}
	}
	\caption{Ablation study for the optimal choice of MDL weight $\lambda_{\text{M}}$.
	Three distinctive motions (two IMOs and the camera motion) exist in these sequences (\emph{Corridor}, \emph{Table}).
	Each column represents a different value of the hyperparameter $\lambda_{\text{M}}$, starting at 1000 and doubling on each right column.
	In general, the smaller the MDL weight, the more motion models are assigned
	(e.g., first three columns). 
	The number of motion models assigned is written below each IWE image.
	A range of endurance, detecting the true number of motions, is found for $\lambda_{\text{M}} \in [8000, 16000]$.}
	\label{fig:ablation study}
\end{figure*}

\textbf{Other Datasets}.
Besides above quantitative results, we also provide an extensive qualitative evaluation on real-world data from EMSMC~\cite{Stoffregen19iccv} dataset, DistSurf~\cite{almatrafi2020distance} dataset, and our own collection.
These sequences cover a wide variety of scenes, ranging from indoor lab environments to outdoor traffic scenes with moving vehicles and pedestrians, including non-rigid motion and HDR scenarios.
We provide exemplary results in Figs.~\ref{fig:emsmc:compare} and~\ref{fig:real:more}.
The first sequence of Fig.~\ref{fig:emsmc:compare} shows a traffic scene captured from above, with three vehicles driving on the road.
The two cars moving to the left have very similar velocities, thus, they are clustered in the same group.
For visual comparison we provide the segmented, motion-compensated images from~\cite{Stoffregen19iccv}.
The second sequence of Fig.~\ref{fig:emsmc:compare} shows an outdoor HDR scene captured with the event camera facing the sun while a pedestrian and a skateboarder pass by.
Our algorithm preserves the motion discrepancy among different parts of the non-rigid human bodies while maintaining the compactness of the segmentation due to the applied MDL term.
The third sequence shows a vehicle passing by in front of buildings.
Our algorithm successfully distinguishes the car from the buildings in the background, which are compactly labeled together, thus identifying the panning camera motion.

In Fig.~\ref{fig:real:more}, the first column shows a fan (whose blades rotate at $\approx$\SI{1800}{\degree/\second})
and a free-falling coin, \ie, a high-speed scenario.
Since our algorithm supports multiple motion models, the fan blades and the coin are successfully detected and distinguished.
The second column shows a traffic scene captured at street level.
The motions of overlapping vehicles can be successfully distinguished.
The third column shows a scene with a pair of waving hands.
The motions of the hand and arm are sometimes segmented from each other when there is a small wrist motion.
The fourth and fifth columns are exemplary results of EVIMO2 dataset, already discussed.
The last column shows another sequence of our own data (different from the one in Fig.~\ref{fig:eyecatcher}), in which a nursing pillow is thrown.
The pillow is detected as an IMO detached from the hand as soon as the apparent in-plane rotation happens.

\subsection{Parameters of the Method}
\label{sec:experim:ablation}

Let us mention how the parameters of the method are set.
First, we process events in packets (\ie, sliding window fashion). 
The number of events $N_e$ may be selected based on the scene dynamics or texture~\cite{Liu18bmvc}.
However, we found that $N_e \in [15000, 30000]$ is a sensible choice for sequences captured by the DAVIS240 or DAVIS346~\cite{Brandli14ssc}.
For sequences acquired by higher resolution sensors (e.g., $640 \times 480$ pixels), $N_e$ is increased proportionally.
As shown in the experiments, motion parameters can be assumed to be constant within each sliding window.

Second, we set the weights $\lambda_{\text{P}} = 40, \lambda_{\text{M}} = 8,000$.
The Potts model weight $\lambda_{\text{P}}$ is set according to the fact that an event typically has at least six spatio-temporal neighbours (At least two edges are linking to the event's pixel location due to the Delaunay triangulation).
The maximum discrepancy for an unary term is 255 according to the IWE negative.
Thus, $\lambda_{\text{P}} = 255 / 6 \approx 40$ so that local consistency would be sacrificed for spatial coherence.
For sequences with relatively small IMOs traversing a textured background, $\lambda_{\text{P}}$ has to be reduced to avoid over-smoothing effects.
The MDL weight $\lambda_{\text{M}}$ is set according to an ablation study, as shown in Fig.~\ref{fig:ablation study}.
Unsurprisingly, we see the resulting number of IMOs becomes smaller as $\lambda_{\text{M}}$ increases.
We observe $\lambda_{\text{M}}$ having a wide range of endurance ($[8000, 16000]$), which makes a good balance between enhancing spatial coherence and circumventing over-smoothing effect.
We find 8,000 a reasonable choice that returns the true number of IMOs in most cases.

Finally, the number of hierarchy levels $N$, which determines the size of the sub-volumes used during initialization, is set empirically.
A good choice (e.g., Fig.~\ref{fig: initialization procedure}) can effectively pick up small IMOs while circumventing cases of bad signal-to-noise ratio (Sec.~\ref{sec:experim:limitations}).
We find that $N=4$ is a good choice for sensors with similar spatial resolution to the DAVIS346, and $N$ may be increased to deal with very small IMOs and/or new devices with higher spatial resolution.

\subsection{Computational Performance}
\label{sec:experim:computational}

Algorithm~\ref{alg: alternating strategy} consists of three main steps:
($i$) initialization, ($ii$) creation of event graph,
and ($iii$) alternating discrete-continuous optimization. 
The initialization is the most time-consuming step.
Using 15,000 events as input and a 4-level subdivision as an example, it takes about \SI{4}{\second} to compute all motion candidates of the model pool.
This time is spent on one motion model type; for multi-model proposals, \eg, $K$ types, the computation is $K$ times larger.
We apply hyper-threading to speed up the process.
Initialization may be expensive for the first set of events, but it can be propagated according to the motion and reutilized for the upcoming events.
The creation of the space-time event graph is efficient, taking \SI{45}{\milli\second}.
The optimization terminates within 3 iterations, and its time is proportional to the size of the motion pool; it takes about \SI{3}{\second}.
The discrete-labeling (graph-cut) sub problem takes the majority of the runtime compared to the negligible time spent on continuous model fitting. 
The proposed method is implemented in C++ on ROS and runs on a laptop with an Intel Core i7-8750H CPU.
The code uses two cores and consumes about 400 MB of memory.

While there is room for improvement in runtime performance to speed up the method by an optimized implementation and/or dedicated hardware, 
event cameras are data-driven sensors that produce more events per second the faster the motion, the smaller the contrast sensitivity and the more texture present in the scene. 
Hence, it is conceivable to modify the above factors to increase the event rate in order to overwhelm any non-trivial event-based method so that it becomes non ``real-time''.
This could be mitigated by decreasing the contrast sensitivity and/or by dropping events (randomly or with some control strategy~\cite{Glover18icra}).
However, this is out of the scope of this work.
In contrast, we think that the most alluring feature of our proposal is the modeling idea of using graph cuts on event-based data, which ($i$) brings well-known principles to the realm of event cameras and ($ii$) allows us to address shortcomings of previous methods, such as the need to either specify the number of clusters in advance or over-segment the scene.

\subsection{Limitations}
\label{sec:experim:limitations}
A limitation observed during initialization entails the size of the IMOs.
Small IMOs, \eg, smaller than $1\%$ of the sensor's spatial resolution (as in the $\emph{multiple objects}$ sequence of~\cite{Mitrokhin18iros}) 
cast signal-to-noise ratio problems during motion proposal generation and are thus unlikely to be properly initialized. 
The size of the IMOs in the other datasets is larger than in the EED dataset, hence detecting IMOs is not an issue in them.

\emph{Noise}. In addition, the performance of our method is affected by the signal-to-noise ratio of the input events because the topological structure of the ST graph deteriorates with increasing event noise (jitter, bursts, etc.). 
To alleviate this issue, a pre-processing de-noising step~\cite{czech2016evaluating, wu2020probabilistic} is suggested. %
We apply a simple de-noising operation, as proposed in~\cite{zhou2020event}, which filters isolated events in the spatio-temporal domain.
To deal with the specific noise in the datasets that is due to ground truth acquisition by the motion capture system (disturbing light from a VICON system's emitters) additional de-noising methods, such as burst filters, are needed.

\section{Conclusion}
\label{sec:conclusion}

We presented a novel method for event-based motion segmentation.
Our approach is a multi-model fitting scheme that jointly clusters events and fits motion models to them.
The proposed graph-based (MRF) formulation with the additional MDL energy term leads to globally consistent and spatially coherent segmentation results using fewest labels.
As a by-product, the method produces labeled, motion-compensated images of warped events that may be used for further processing (\eg, recognition).
A thorough evaluation demonstrated the versatility of our method in scenes with different motion patterns and unknown number of independent motions.
We also showed that the method is able to bring the advantages of event-based cameras to tackle traditionally difficult scenarios for standard frame-based cameras, such as segmentation of fast moving objects (that would cause motion blur) or in HDR conditions.
Finally, we hope this work inspires new research in the topic of segmentation with event-based cameras, a paramount but rather unexplored topic.

\section{Acknowledgement}
We thank Dr. Timo Stoffregen and his co-authors for providing test data from~\cite{Stoffregen19iccv}.
We thank Chethan Parameshwara for providing the comparison images in Fig.~\ref{fig: qualtitative evaluation EVIMO}.
We thank Chuhao Liu and Hao Xu for their help during data collection.


\vskip -2\baselineskip plus -1fil

\begin{IEEEbiography}[{\includegraphics[width=1in,height=1.25in,clip,keepaspectratio]{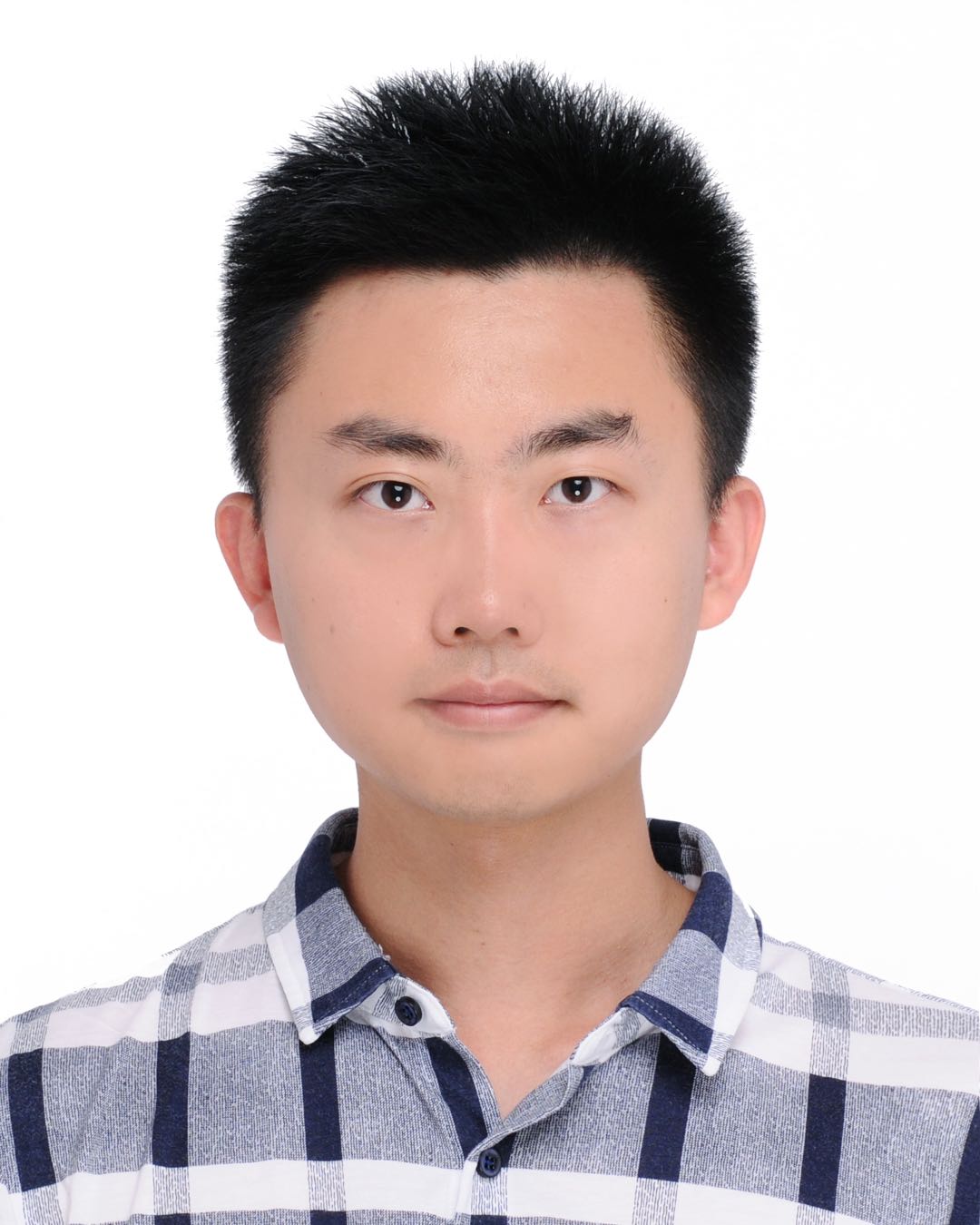}}]{Yi Zhou} received the B.Sc. degree in aircraft manufacturing and engineering from Beijing University of Aeronautics and Astronautics, Beijing, China in 2012, and the Ph.D. degree in engineering and computer science from the Australian National University, Canberra, Australia in 2018.
From 2019 to 2021 he was a postdoctoral research fellow at the Hong Kong University of Science and Technology, Hong Kong SAR, China.
He joined the School of Robotics at Hunan University in October 2021 as an Associate Professor.
His research interests include visual odometry / simultaneous localization and mapping, geometry problems in computer vision, and dynamic vision sensors.
He was awarded the NCCR Fellowship Award for the research on event based vision in 2017 by the Swiss National Science Foundation through the National Center of Competence in Research Robotics.
\end{IEEEbiography}

\begin{IEEEbiography}[{\includegraphics[width=1in,height=1.25in,clip,keepaspectratio]{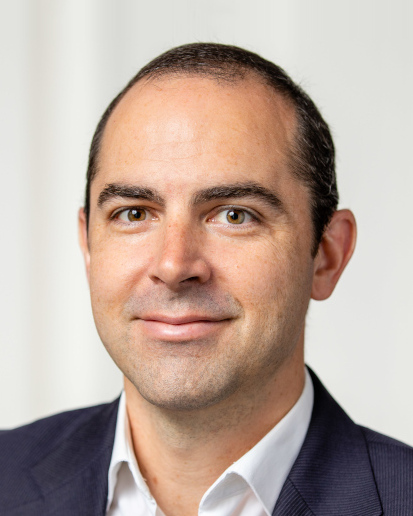}}]{Guillermo Gallego} (SM'19) is Associate Professor at Technische Universit\"at Berlin, Berlin, Germany, in the Dept. of Electrical Engineering and Computer Science, and at the Einstein Center Digital Future, Berlin, Germany.
He is also a Principal Investigator at the Science of Intelligence Excellence Cluster, Berlin, Germany.
He received the PhD degree in Electrical and Computer Engineering from the Georgia Institute of Technology, USA, in 2011, supported by a Fulbright Scholarship.
From 2011 to 2014 he was a Marie Curie researcher with Universidad Politecnica de Madrid, Spain, and from 2014 to 2019 he was a postdoctoral researcher at the Robotics and Perception Group, University of Zurich and ETH Zurich, Switzerland.
His research interests include robotics, computer vision, signal processing, optimization and geometry. 
\end{IEEEbiography}

\begin{IEEEbiography}[{\includegraphics[width=1in,height=1.25in,clip,keepaspectratio]{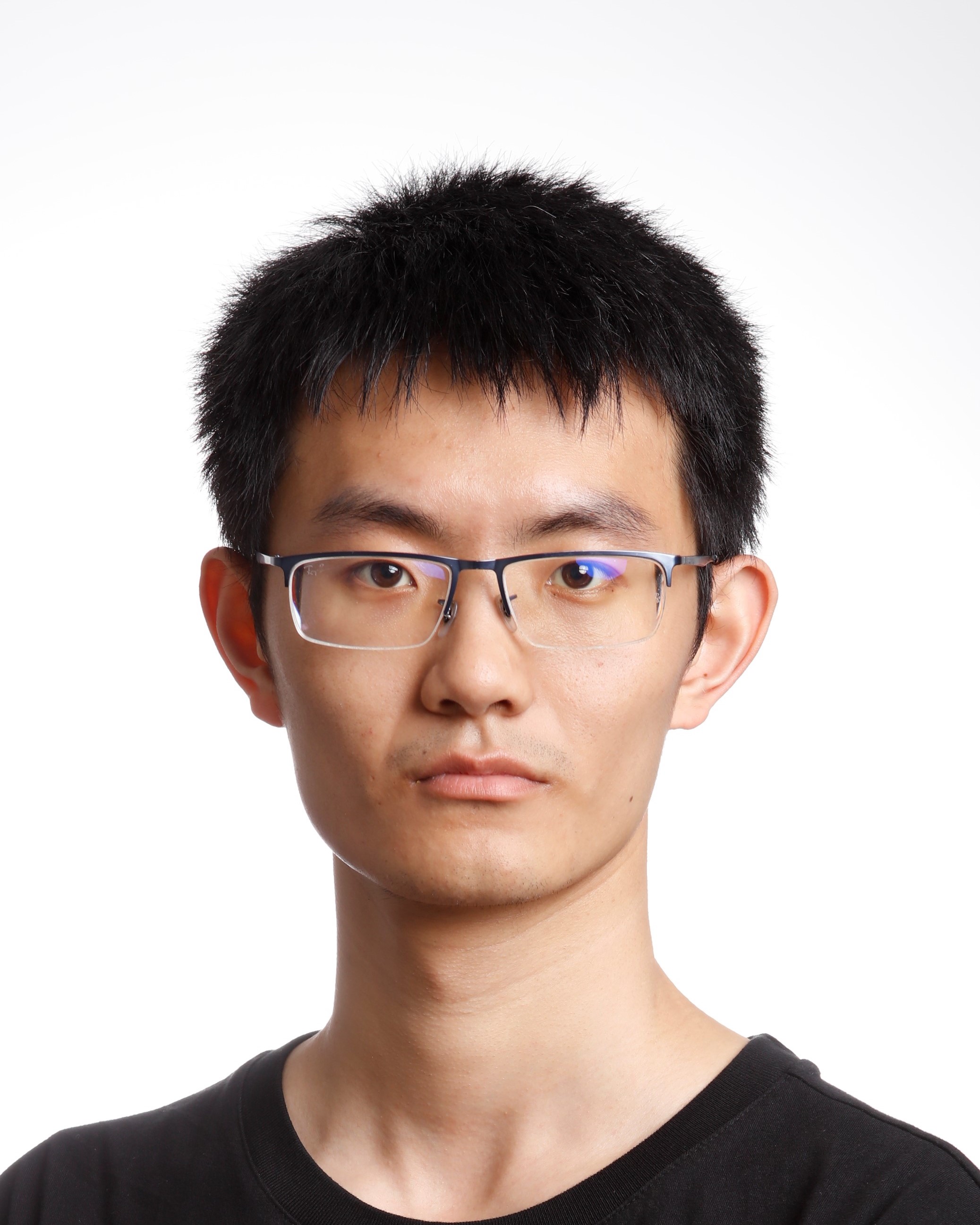}}]{Xiuyuan Lu}
received the B.Eng. degree in computer science from the Hong Kong University of Science and Technology (HKUST), Hong Kong, in 2020. 
He is currently a PhD candidate in electronic and computer engineering at the Hong Kong University of Science and Technology, Hong Kong.
His research interests include event-based vision and visual odometry/simultaneous localization and mapping. 
\end{IEEEbiography}

\begin{IEEEbiography}[{\includegraphics[width=1in,height=1.25in,clip,keepaspectratio]{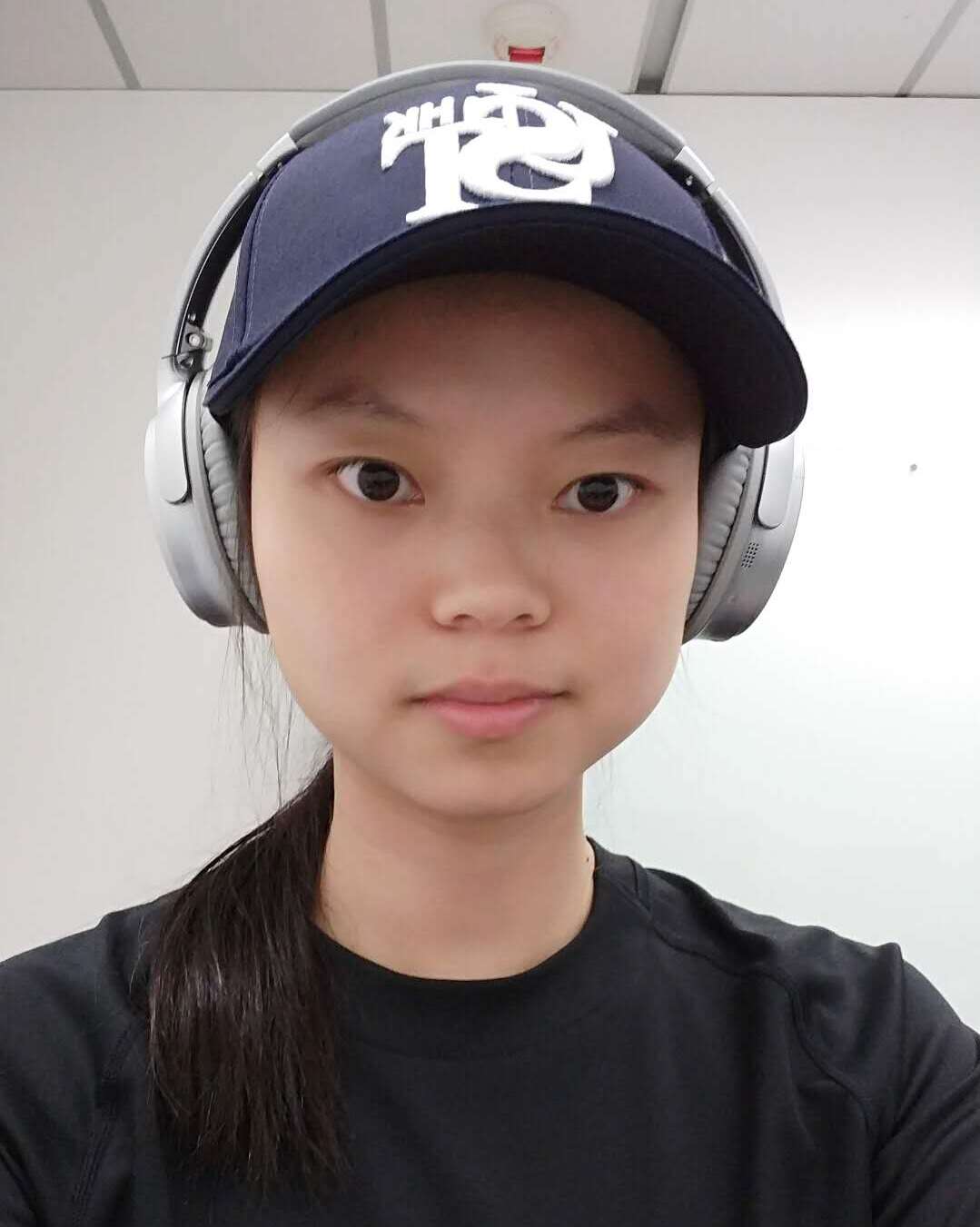}}]{Siqi Liu}
received the B.Eng and M.Eng degree in control science and engineering from the Harbin Engineering University. 
She is currently working towards a degree in robotics with the Hong Kong University of Science and Technology (HKUST).
Her research interests include state estimation, sensor fusion, object detection, flexible baseline stereo, and event based feature tracking.
\end{IEEEbiography}

\begin{IEEEbiography}[{\includegraphics[width=1in,height=1.25in,clip,keepaspectratio]{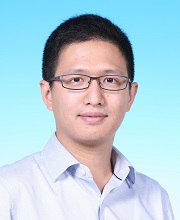}}]{Shaojie Shen} (Member, IEEE) received his B.Eng. degree in Electronic Engineering from the Hong Kong University of Science and Technology (HKUST) in 2009. 
He received his M.S. in Robotics and Ph.D. in Electrical and Systems Engineering in 2011 and 2014, respectively, all from the University of Pennsylvania. 
He joined the Department of Electronic and Computer Engineering at the HKUST in September 2014 as an Assistant Professor, and is promoted to Associate Professor in July 2020. 
He is the founding director of the HKUST-DJI Joint Innovation Laboratory. His research interests are in the areas of robotics and unmanned aerial vehicles, with focus on state estimation, sensor fusion, localization and mapping, and autonomous navigation in complex environments. He was the regional program chair of SSRR 2017 and program co-chair of SSRR 2015. 
He is currently serving as associate editor for T-RO, and senior editor for IROS 2020-2022. He and his research team received Honorable Mention status for the IEEE T-RO Best Paper Award in 2020 and 2018, and won the Best Student Paper Award in IROS 2018, Best Service Robotics Paper Finalist in ICRA 2017, Best Paper Finalist in ICRA 2011, and Best Paper Awards in SSRR 2016 and SSRR 2015.
\end{IEEEbiography}

\end{document}